\documentclass[review]{elsarticle}

\usepackage{lineno,hyperref}
\modulolinenumbers[5]

\usepackage{graphicx}
\usepackage{amssymb}
\usepackage{amsmath}
\usepackage{tabularx}
\usepackage{lscape}
\usepackage{ltablex}
\usepackage{url}
\usepackage{color}
\usepackage{type1cm}
\usepackage {colortbl,array,xcolor}

\usepackage[whole]{bxcjkjatype} 

\newcommand{\II}{\rm{I\hspace{-.1em}I }}
\newcommand{\III}{\rm{I\hspace{-.1em}I\hspace{-.1em}I }}

\usepackage{color}    
\usepackage{umoline} 
\usepackage{proofread} %
\noproofreadmark 

\usepackage[section,numberedsection=autolabel,nopostdot,toc,
 acronym,
 nomain,
 nonumberlist
]{glossaries}

\makeglossaries

\setacronymstyle{long-short}

\newacronym[type=\glsdefaulttype]{HF-PGM}{HF-PGM}{hippocampal formation-inspired probabilistic generative model}
\newacronym[type=\glsdefaulttype]{HF}{HF}{hippocampal formation}
\newacronym[type=\glsdefaulttype]{SLAM}{SLAM}{simultaneous localization and mapping}
\newacronym[type=\glsdefaulttype]{PGM}{PGM}{probabilistic generative model}
\newacronym[type=\glsdefaulttype]{MEC}{MEC}{medial entorhinal cortex}
\newacronym[type=\glsdefaulttype]{AI}{AI}{artificial intelligence}
\newacronym[type=\glsdefaulttype]{SCID}{SCID}{structure-constrained interface decomposition}
\newacronym[type=\glsdefaulttype]{SERKET}{SERKET}{symbol emergence in the robotics tool kit}
\newacronym[type=\glsdefaulttype]{CA}{CA}{cornu ammonis area}
\newacronym[type=\glsdefaulttype]{DG}{DG}{dentate gyrus}
\newacronym[type=\glsdefaulttype]{LEC}{LEC}{lateral entorhinal cortex}
\newacronym[type=\glsdefaulttype]{Sb}{Sb}{subiculum}
\newacronym[type=\glsdefaulttype]{ParaSb}{ParaSb}{parasubiculum}
\newacronym[type=\glsdefaulttype]{PreSb}{PreSb}{presubiculum}
\newacronym[type=\glsdefaulttype]{PER}{PER}{perirhinal cortex}
\newacronym[type=\glsdefaulttype]{POR}{POR}{postrhinal cortex}
\newacronym[type=\glsdefaulttype]{RSC}{RSC}{retrosplenial cortex}
\newacronym[type=\glsdefaulttype]{BRA}{BRA}{brain reference architecture}
\newacronym[type=\glsdefaulttype]{BIF}{BIF}{brain information flow}
\newacronym[type=\glsdefaulttype]{HCD}{HCD}{hypothetical component diagram}
\newacronym[type=\glsdefaulttype]{SpCoSLAM}{SpCoSLAM}{online spatial concept acquisition with simultaneous localization and mapping}
\newacronym[type=\glsdefaulttype]{LSTM}{LSTM}{long short-term memory}
\newacronym[type=\glsdefaulttype]{RNN}{RNN}{recurrent neural network}
\glsdisablehyper

\journal{Neural Networks}

\begin{document}

\begin{frontmatter}

\title{
Hippocampal formation-inspired probabilistic generative model 
}

\author[Ritsumei]{Akira Taniguchi\corref{cor}}
\ead{a.taniguchi@em.ci.ritsumei.ac.jp}

\author[WBAI]{Ayako Fukawa}
\ead{fukawa@yairilab.net}

\author[WBAI,UT,RIKEN]{Hiroshi Yamakawa}
\ead{ymkw@wba-initiative.org}

\cortext[cor]{Corresponding author}
\address[Ritsumei]{Ritsumeikan Univervsity, 1-1-1 Noji-Higashi, Kusatsu, Shiga 525-8577, Japan}
\address[WBAI]{The Whole Brain Architecture Initiative, Nishikoiwa 2-19-21, Edogawa-ku, Tokyo, 133-0057, Japan}%
\address[UT]{The University of Tokyo, 7-3-1 Hongo, Bunkyo-ku, Tokyo 113-0033, Japan}
\address[RIKEN]{RIKEN, 6-2-3, Furuedai, Suita, Osaka 565-0874, Japan}

\begin{abstract}
\addspan{
In building artificial intelligence (AI) agents, referring to how brains function in real environments can accelerate development by reducing the design space.
In this study, we propose a probabilistic generative model (PGM) for navigation in uncertain environments by integrating the neuroscientific knowledge of hippocampal formation (HF) and the engineering knowledge in robotics and AI, namely, simultaneous localization and mapping (SLAM). 
We follow the approach of brain reference architecture (BRA)~\citep{Yamakawa2021-yy} to compose the PGM and outline how to verify the model.
To this end, we survey and discuss the relationship between the HF findings and SLAM models.
The proposed \textit{hippocampal formation-inspired probabilistic generative model} (HF-PGM) is designed to be highly consistent with the anatomical structure and functions of the HF.
By referencing the brain, we elaborate on the importance of integration of egocentric/allocentric information from the entorhinal cortex to the hippocampus and the use of discrete-event queues. 
}

\end{abstract}
 
\end{frontmatter}


\section{Introduction}
\label{sec:introduction} 

%
\addspan{In building artificial intelligence (AI) agents, referring to how brains function in real environments can accelerate development by reducing the design space.
}
Hippocampal formation (HF) supports crucial neural capabilities, such as spatial cognition, self-localization for navigation, mapping, and episodic memory.
In neuroscience, HF and its functions have attracted increasing attention in recent years.
The hippocampus has long been considered the brain region responsible for configuring the cognitive map \citep{Tolman1948, Okeefe1978placecells}.
To this end, designated neurons, such as place cells in the hippocampus~\citep{Okeefe1978placecells} and grid cells in the medial entorhinal cortex (MEC), exist to execute these functions~\citep{Fyhn2004,Hafting2005gridcells}.
From the perspective of computational neuroscience, numerous computational model-based studies have focused on functions involving the hippocampus~\citep{milford2004ratslam, Madl2015, Schapiro2017, Banino2018, Kowadlo2019, Scleidorovich2020}.
Alongside these computational studies, the use of brain-inspired AI and intelligent robotics is crucial to the implementation of these spatial functions.
From an engineering perspective, simultaneous localization and mapping (SLAM)~\citep{thrun2005probabilistic} represents a typical approach in computational geometry and robotics.
Spatial cognition and place understanding are important challenges that must be overcome to facilitate the advance of  robotics~\citep{taniguchi2018TCDSsurvey}.
However, despite the abundancy of neuroscience knowledge related to HF and the progress in AI technology, combining knowledge from both fields and applying it to robotics remains a major challenge.

\textbf{Purposes}:
This study aims to bridge the gap between neuroanatomical/biological findings of the HF and engineering technologies of probabilistic generative models (PGMs), particularly in AI and robotics.
{This paper is a feasibility study on the methodology proposed by \citet{Yamakawa2021-yy}.}
We establish a correspondence between the function/structure of the HF in neuroscience and spatial cognitive methods in robotics.
The main objectives of this study are as follows.
\begin{itemize}
 \item To provide suggestions for the construction of a computational model with functions of HF by surveying the association between SLAM in robotics and HF in neuroscience.
 \item To construct a \textit{brain reference architecture} (BRA) that operates with biologically valid and consistent functions, as a specification for implementing a brain-inspired model.
\end{itemize}

\textbf{Type of paper}:
This report is a hypothesis-suggestion paper that presents a novel argument, interpretation, or model intended to introduce a hypothesis/theory, based on a literature review, and provides the direction for its verification.
We construct a \textit{hippocampal formation-inspired probabilistic generative model} (HF-PGM) as a highly adequate hypothesis by referencing neuroanatomical/biological findings.
{The adequacy of the constructed model can be evaluated} by adopting the \textit{structure-constrained interface decomposition} (SCID) method~\citep{Yamakawa2021-yy} for hypothesis generation.
A \textit{generation-inference process allocation} task is used to solve particular problems related to the mapping of PGMs to brain circuits. 
This task involves allocating all anatomical connections to either the generative or the inference process.
Here, we learn the connections between modules from the brain and utilize engineering technologies for the parts having insufficient findings of the brain structure and function.
The available findings on HF have been primarily gathered from studies on rodents and only partially from the human brain.
Engineering technologies refer to methods related to spatial awareness (e.g., SLAM, navigation, place recognition, spatial concept formation, and semantic mapping)~\citep{thrun2005probabilistic,kostavelis2015semantic}.
Thus, we present computational HF models to support a feasible hypothesis from an engineering perspective.
Furthermore, we provide suggestions for the construction of methods to study spatial cognitive functions using robotics and {AI} technology from the perspective of neuroscience.
The proposed PGM is designed using an extension of the development method proposed by \citet{Yamakawa2021-yy}. 
{Therefore, the adequacy of the architecture is based on the evaluation criteria proposed in the same work.}
These evaluation criteria require (a) the brain information flow (BIF) to be consistent with the anatomical findings, (b) the hypothetical component diagram (HCD) to be consistent with the structure of the BIF, and (c) the HCD to be able to achieve the expected computational functions. 
The proposed PGM corresponding to the neural circuitry of the HF is obtained through a design that is a natural extension of SLAM based on engineering practice, and the design procedure is detailed in Section~\ref{sec:hpf-pgm:model}. 
Hence, the feasibility (c) of HCD is computationally {supported} without any new implementations or simulations.

\textbf{Contributions}:
The main contributions of this study are as follows.
\begin{itemize}
 \item We proffer and apply the generation-inference process allocation, an approach that allows neural circuits to be interpreted as PGM, for the first time.
 \item We clarify that the function and structure of HF can be consistently represented as an extension of the previously reported SLAM models by considering relevant findings on HF and SLAM.
 \item We show that the phase precession of brain activity in the HF can be formulated as a \textit{discrete-event queue} by repeatedly performing sequential Bayesian inference on the joint probability distribution of the variables.
 \item We illuminate the direction of future challenges that are important for the development of HF-inspired models through discussions on spatial movement, hierarchies, and physical constraints. 
\end{itemize}

This paper is structured as follows. 
Section~\ref{sec:background} describes the background and motivation.
Section~{\ref{sec:bra}} describes the methodology for constructing a brain reference architecture and PGM.
Section~\ref{sec:neuroscience} summarizes the neuroscientific findings on HF.
Section~\ref{sec:hpf-slam} describes relevant topics, including brain-inspired SLAM, computational models of the HF, concept formation and semantic understanding of location and space, and deep generative models. 
Section~\ref{sec:hpf-pgm} describes the proposed HF-PGM from the perspective of the model structure. 
We construct a novel PGM that is highly consistently with the neuroanatomical/biological findings of HF by integrating allocentric/egocentric visual information.
{Section~\ref{sec:hpf-pgm:queue} describes the formulation of a discrete-event queue based on the phase precession queue assumption in HF as one of the inference algorithms for PGMs.}
Section~\ref{sec:conclusion} summarizes the findings of this study and discusses future directions.

\section{Backgrounds and motivation}
\label{sec:background}

\subsection{Bridging the gap between two fields}
\label{sec:background:gap}

Models facilitated by engineering approaches, such as robotics and AI, are capable of supporting HF modeling. 
Numerous aspects of the hippocampus remain unknown. 
Hence, there are limitations to creating an HF model solely based on existing hippocampus-related knowledge in neuroscience and biology. 
Models for problem-solving based on engineering do not encounter these types of restrictions.
It is possible to build a model with functions similar to those of the hippocampus without actively utilizing its knowledge.
Such engineering models constructed for solving actual tasks may have implications for neuroscience and biology.

Furthermore, the knowledge on neuroscience is useful for engineering applications in robotics and AI.
Robotics generally depend on a particular task and involves practical applications (e.g., accurate 3-dimensional (3D) modeling of the environment. 
Brain-inspired {AI} emphasizes learning from the brain and mapping brain circuitry and functions, followed by searching for possible practical applications.
Building a system that refers to the brain that actually operates with functions for various tasks opens engineering possibilities from novel perspectives.

Brain reference architecture, which links neuroscience and robotics while enabling their application in robots, is currently an important field of research.

\subsection{PGM-based cognitive architecture}
\label{sec:background:pgm}

Probabilistic models that have succeeded as neocortical computational models are also useful for modeling HF.
In particular, PGMs represent the process of generating observational stimuli by assuming dependencies between random variables. 
Probabilistic inference (e.g., Bayesian) may be employed by adopting PGMs, which infer states behind sensory stimuli as latent variables, enabling the construction of internal representations.
The observed variables correspond to the stimuli. 
Adopting PGMs is valid from the perspective of the Bayesian brain hypothesis, which states, ``the brain represents sensory signals probabilistically in the form of probability distributions''~\citep{Knill2004,doya2007bayesian}. 
Furthermore, a reason for modeling brain-inspired architecture using PGMs is that the Neuro-symbol emergence in the robotics tool kit (SERKET) architecture can be used both theoretically and practically~\citep{Taniguchi2020neuro}. 
Neuro-SERKET enables PGMs having several functions to have distributed development, which can be used to develop integration modules. 
This architecture enables the integration of models that imitate other regions in the brain into a whole-brain model. 
Therefore, the PGMs for HF proposed in this study may be used as a module to represent the whole-brain's integrated cognitive architecture~\citep{Taniguchi2021wb-pgm}.

\begin{figure}[!tb]
    \centering
    \includegraphics[width=1.0\textwidth]{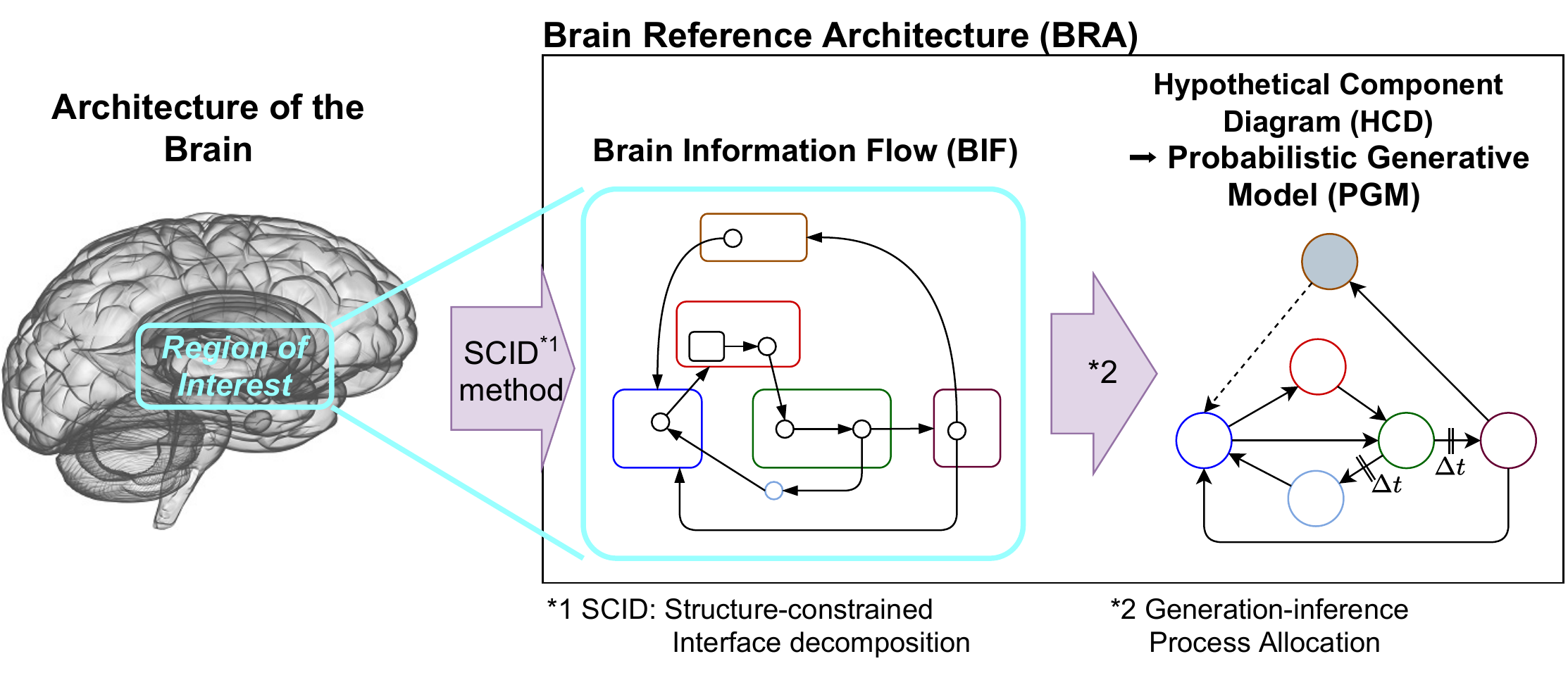}
    \caption{
        Overview of modeling processes for brain reference architecture \addspan{(BRA)}. 
        The brain information flow \addspan{(BIF)} and hypothetical component diagrams \addspan{(HCD)} are constructed using the structure-constrained interface decomposition \addspan{(SCID)} method.
        The probabilistic generative model (PGM) is one of the expression formats of the hypothetical component diagrams constructed by generation-inference process allocation.
    }
    \label{fig:scid_gipa}
\end{figure}

\section{Brain reference architecture ({BRA}) construction}
\label{sec:bra}

This section describes the methodology for constructing the PGM as a BRA.
Figure~\ref{fig:scid_gipa} shows the relationship of components for the modeling processes.
Section~\ref{sec:background:wba} introduces an overview on BRA.
Section~\ref{sec:hpf-pgm:scid} {describes construction process and adequacy evaluation for BIF and HCD.}
Section~\ref{sec:hpf-pgm:gipa} describes a problem-solving approach for associating anatomical structures in the brain with PGMs (i.e., generation-inference process allocation).
The generation-inference process allocation is devised and demonstrated for the first time in this study.

\subsection{Brain reference architecture}
\label{sec:background:wba}

This study adopts an approach for building parts of modules in a whole-brain architecture, which is a brain-inspired artificial general intelligence development approach that emphasizes the architecture. 
%
Design methodologies for combining machine learning that imitates the mesoscopic-level brain structures of rodents and humans are being studied to gather knowledge on the architecture of the brain. 
Specifically, the whole-brain architecture initiative standardizes information corresponding to the fulfilment of requirement specifications for brain-inspired software as BRA data\footnote{{\url{https://wba-initiative.org/wiki/en/brain_reference_architecture}}} and promotes role sharing in the BRA design and utilization. 
The whole-brain architecture initiative includes a manual\footnote{{BRA Data Preparation Manual: \url{https://docs.google.com/document/d/1_t3W_dkFmGjfBhz3_EEZ2FCYyzrJi_1ZOtoPlOru8dc/edit\#heading=h.6766fia4kgtf}}} for preparing brain reference architecture (BRA) as the standard notation for guiding the development of brain-inspired software.

BRA, which is the reference architecture at the mesoscopic level, represents a consistent description of (i) the brain information flow (BIF) related to anatomical structures and (ii) the hypothetical component diagram (HCD) related to its computational functions \citep{Yamakawa2021-yy}. 
BIF is a directed graph comprising partial circuits and connections that represent the anatomical structure of neural circuits in the brain. 
HCD is a directed graph describing component dependencies, which can be associated with any BIF sub-graph. 
The functional mechanisms described in HCD are hypothetical, as the name implies, and different hypotheses may be presented by neuroscientists from different perspectives. 
In many cases, brain-inspired software is implemented by engineers who do not necessarily have a deep understanding of neuroscience. 
Therefore, it is useful to examine HCD candidates that cannot be clearly dismissed based on current neuroscience knowledge, even if they cannot be confirmed as ground truth.

\subsection{{Construction process and adequacy evaluation}}
\label{sec:hpf-pgm:scid}

In this study, we use a \textit{structure-constrained interface decomposition} (SCID) method developed by the whole-brain architecture initiative~\citep{Yamakawa2021-yy} to design an HCD consistent with the anatomical structure of the HF.
Using the SCID method, \citet{Fukawa2020} applied the concept to the neural circuit of HF for the first time and identified a circuit that performs a path integration function on MEC.


\begin{figure}[!tb]
    \centering
    \includegraphics[width=0.98\textwidth]{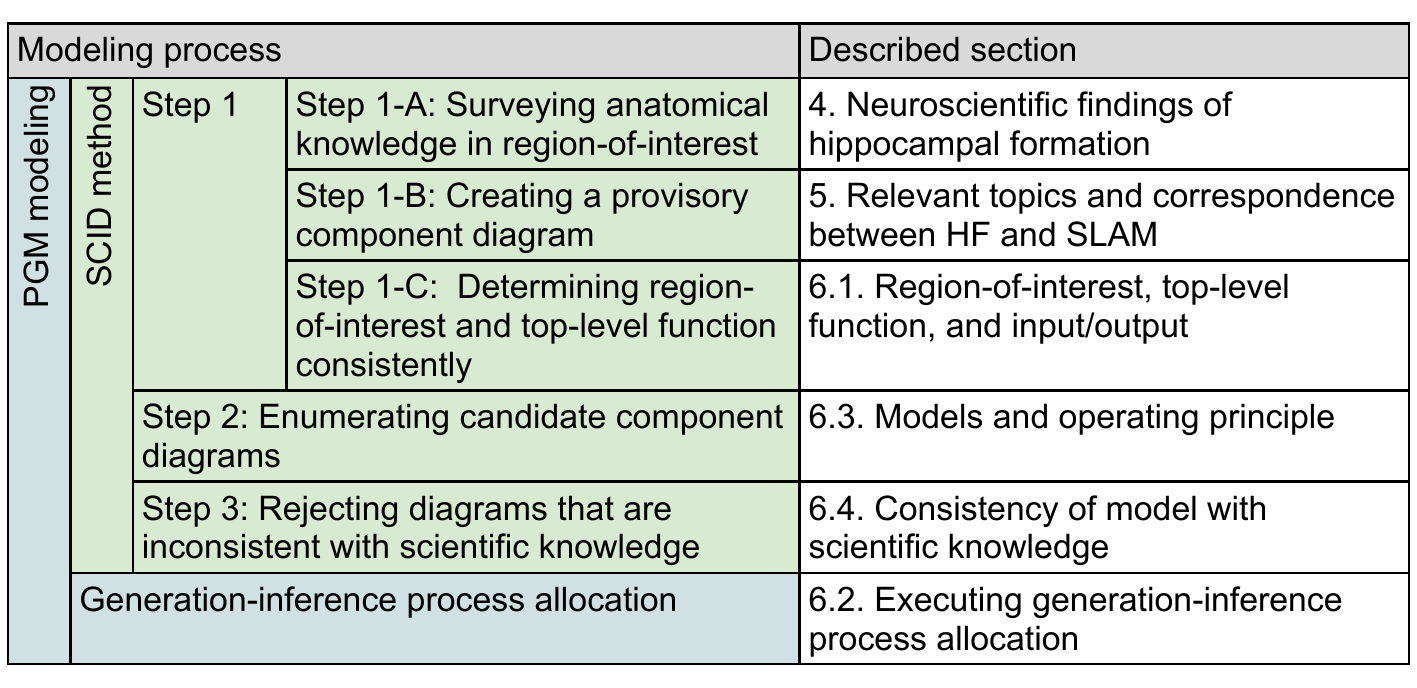}
    \caption{
    {Modeling processes and section structure by structure-constrained interface decomposition (SCID) method and generation-inference process allocation}
    }
    \label{fig:model_process}
\end{figure}

We created one plausible HCD by applying steps 1 and 2 {(see \ref{apdx:sec:bra} for the detail)} mainly to the HF.
{In this study, PGM is positioned as a form of expression for HCD.}
Figure~\ref{fig:model_process} shows a table of PGM modeling processes in the SCID method.
The execution result of step 1-A is mainly explained in Section~\ref{sec:neuroscience}.
BIF is shown in Fig.~\ref{fig:hpf_circuit}.
The execution result of step 1-B is explained in Section~\ref{sec:hpf-slam}.
The execution result of step 1-C is explained in Section~\ref{sec:hpf-pgm:roi}.
The execution result of Step 2 is explained in Section~\ref{sec:hpf-pgm:model}.
The contents partially examined in Step 3 are described in Section~\ref{sec:hpf-pgm:step3}.

The authenticity of BIF was established by \citet{Yamakawa2021-yy} in the section ``Adequacy evaluation of BIF''. 
We further elaborate on this adequacy in Section~\ref{sec:neuroscience}. 
Similarly, the consistency of HCD with BIF has been described in the section ``Adequacy evaluation of HCD'' in \citet{Yamakawa2021-yy}. 
The functionality of HCD is ensured by the presence of SLAM models (details are presented in Section~\ref{sec:hpf-pgm:model}. 
The model was created to be consistent with BIF, thus ensuring structure-consistency (discussed in Section~\ref{sec:hpf-pgm:step3}).

\subsection{Generation-inference process allocation}
\label{sec:hpf-pgm:gipa}

We describe the task developed in this study to solve particular problems for mapping brain circuits when modeling PGMs. 
PGMs have the restriction that they must be directed acyclic graphs; hence, loops cannot be represented.
(See \ref{apdx:pgm_slam:PGM} for a basic description of PGMs.)
However, some brain circuits have loop structures.
In most cases, it is difficult to assign acyclic PGMs to a brain circuit.
Furthermore, in ordinary PGMs, signal propagation in the inference process causes signals to propagate in the direction opposite to that of the links used during the generative process.
In contrast, in brain neural circuits, signal propagations between regions propagated by electrical spikes that propagate terminally on axons are essentially unidirectional. 
Generally, modeling a PGM of any existing interarea connection in the brain is a difficult task.

To eliminate this restriction, we adopt a combination of generative/inference models\footnote{As other solutions, we can use probabilistic models based on factor/undirected graphs instead of generative models.}.
In other words, we assume that an \textit{amortized variational inference} is introduced, as described in \ref{apdx:pgm_slam:PGM}.
The amortized variational inference can define the link structure of the inference process without depending on the link structure of the generative process.
The model used for amortized variational inference can be designed with a high degree of freedom, as long as it is consistent with the link structure of the generative process. 
Even if loops are present, there is a time delay in the signal transduction in the neural circuits of the brain.
Therefore, we introduce \textit{next-time generation}\footnote{It means assuming state-space models for time-series data.}, which expresses the progress of time for the generation process, as shown in Fig.~\ref{fig:slam_gm} (b).
Therefore, it becomes easy to relate the link structure of PGMs to the structure of actual brain neural circuits.

This allocating task is the generation-inference process allocation.
This task consistently allocates all links into a generative or inference model in a dependency network.
In neural circuits adjacent to the cortex, attention must be paid to avoid inconsistencies at the cortex interface.
If the neuroscientific findings are uncertain, a reasonable and feasible engineering model is chosen.
As a prerequisite for performing this task, each node of these variables\footnote{Nodes are basically random variables; furthermore, they can be represented as temporary variables during the calculation process of deterministic variables.} is principally associated with a uniform circuit (see \cite{Yamakawa2021-yy}), which is the minimum descriptive unit of the BIF.

Generally, applying generation-inference process allocation to any existing interarea connections in the brain is not an easy task. 
However, the major interarea connections of the neocortex can be allocated to either the generative or inference process.
In the neocortex, a feedforward pathway transmits signals from lower to higher areas while processing signals received by sensors, and a feedback pathway transmits signals in the opposite direction \citep{Markov2013-zd,Markov2014-ez}.
In computational neuroscience theories (e.g., Bayesian brain \citep{Knill2004,doya2007bayesian} and predictive coding \citep{Rao1999}), inference and generation are assumed to be processed by the feedforward and feedback pathways, respectively.
Therefore, the flow of the inference model, represented by the dotted arrow in the graphical model, is associated with the pathway of the feedforward signal, and the flow of the generative model represented by the solid arrow is associated with the pathway of the feedback signal.

\section{Neuroscientific findings of hippocampal formation}
\label{sec:neuroscience}

In this section, we survey the connections and functions of HF and its surrounding areas. 
The core region of interest of the HF focused in this study includes the cornu ammonis-1 and -3 areas (CA1 and CA3) and dentate gyrus (DG) in the hippocampus, lateral/medial entorhinal cortex (LEC/MEC), subiculum (Sb), and parasubiculum (ParaSb).
The presubiculum (PreSb), perirhinal cortex (PER), postrhinal cortex (POR), retrosplenial cortex (RSC), and medial septum are also investigated as areas adjacent to the HF, that is, as regions for sending and receiving the signal.

\subsection{Anatomical and physiological findings}
\label{sec:neuroscience:connections}

\begin{figure}[!tb]
    \centering
    \includegraphics[width=1.0\textwidth]{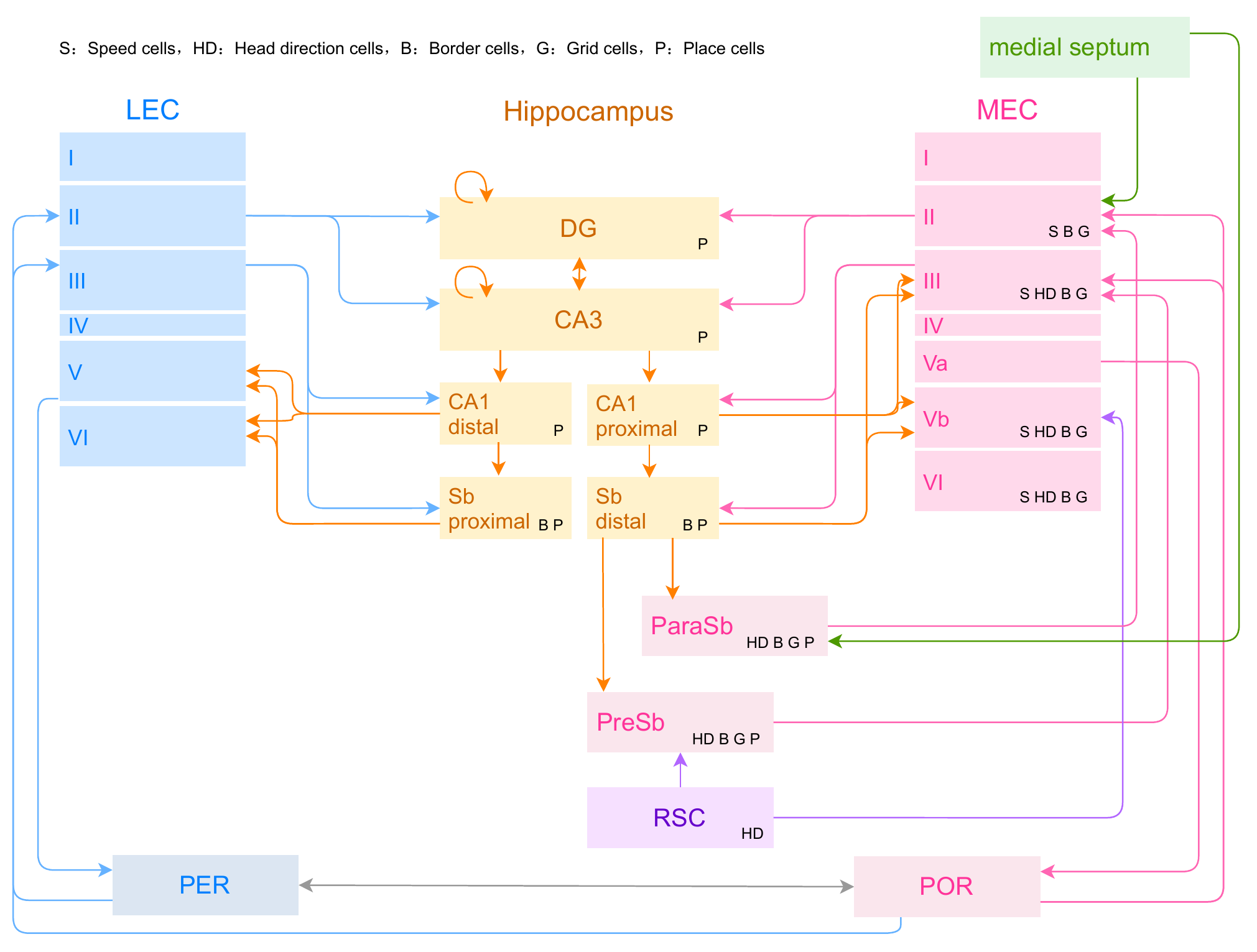}
    \caption{
        Brain information flow (BIF) represents the connection of neural circuits between HF and its surrounding areas. 
        It is based on anatomical studies and reviews \citep{Amaral1989,Witter2000,Fukawa2020}.
        {Blue, lateral entorhinal cortex (LEC) and perirhinal cortex (PER); red, medial entorhinal cortex (MEC), presubiculum (PreSb), parasubiculum (ParaSb) and postrhinal cortex (POR); yellow, the hippocampus and subiculum (Sb); green, the medial septum; and purple, retrosplenial cortex (RSC).}
        {The hippocampus includes the cornu ammonis-1 and -3 areas (CA1 and CA3) and dentate gyrus (DG).}
        Colored arrows indicate that the signal is sent from the area indicated by that color to the other area at the tip of the arrow. 
    }
    \label{fig:hpf_circuit}
\end{figure}

With respect to neuroscientific findings, we conducted a survey by adding detailed findings on LEC and hippocampus to the findings in the review of \citet{Fukawa2020} centered on the hippocampus and MEC.
These findings are also based on an original anatomical review of the structure of HF~\citep{Amaral1989,Witter2000}.
Figure~\ref{fig:hpf_circuit} shows the connection relationship of the HF circuit, which includes the hippocampus, Sb, PreSb, ParaSb, and entorhinal cortex. 
The hippocampus comprises DG, CA1, and CA3.
The entorhinal cortex is divided into MEC and LEC.
CA1 and Sb are further separated distally and proximally~\citep{Knierim2014}.
In this study, MEC is divided into six layers: I, II,  III, Va, Vb, and VI, and LEC is similarly divided into six layers: I, II, III, IV, V, and VI~\citep{Shepherd2013, Fukawa2020}. 
MEC IV is excluded because it has few neurons.
As input to LEC, a connection exists from the POR and PER to LEC II and III~\citep{Nilssen2019}.
The output from LEC has a connection from LEC II to DG and CA3 and from LEC III to CA1 and Sb.
From CA3, there are two connections, Sb proximal through CA1 distal and Sb distal through CA1 proximal~\citep{Knierim2014}.
The nonspatial signal is conveyed from LEC to the hippocampus~\citep{Hargreaves2005}.
The MEC and LEC signals are integrated by the DG and CA3 in the hippocampus~\citep{Chen2013, Knierim2014}.

Neural cells with various expressions have been observed in the hippocampus and its surrounding areas.
Place cells exist in CA3 and CA1~\citep{Okeefe1978placecells}; they are active only when an animal enters a specific place in the environment; they do not fire elsewhere.
It is thought that an environmental cognitive map is stored as a neural circuit according to the place cells~\citep{Okeefe1978placecells}.
Grid cells exist in MEC, ParaSb, and PreSb~\citep{Hafting2005gridcells}.
Although the activity of place cells represents one place in the environment, grid cells are activated in a hexagonal grid at multiple places in the environment.
Head-direction cells exist in MEC, PreSb, ParaSb, and RSC~\citep{Taube1990, Taube2007, Grieves2017}; they are active only when the head faces a specific direction, regardless of the location of the viewer.
Border cells and boundary vector cells are cells in Sb and MEC that selectively fire near the border, irrespective of whether there are objects or obstacles~\citep{lever2009}.
Speed cells exist in MEC and exhibit activity that depends on the speed of movement of the subject~\citep{Kropff2015, Hinman2016}.
Spatial view cells that respond to the current landscape are known to exist on the primate's hippocampus~\citep{Rolls2013}.
Furthermore, event cells have been discovered that represented the content and order of the events experienced~\citep{Terada2017}.
Events that are not evenly spaced in time are processed centrally by the LEC and stored only when the state changes.

\subsection{Functions}
\label{sec:neuroscience:functions}

HF is a brain region that controls short-term and episodic memory in vertebrates. It is deeply involved in spatial memory functions, such as spatial learning and exploration.
The behavior of an animal when traveling to a destination by selecting an appropriate route while simultaneously acquiring environmental information is called navigation.
Cognitive mapping, where animals form a map of the spatial positional relationship of various items in the environment by exploration and act accordingly using this map, is proposed as a psychological concept deeply related to navigation~\citep{Tolman1948,Giocomo2011}.
The hippocampus is thought to be crucial to the formation of cognitive maps in mammals.
	
Self-localization, where animals consistently recognize their current position, is indispensable for the navigation and formation of cognitive maps.
One of the functions that support the self-localization ability of animals is path integration~\citep{McNaughton2006}.
Path integration is a function that outputs self-position after movement upon input of the initial position, head-direction, and movement signals, including speed and movement direction \citep{Raudies2015a}.
These researchers argued that the region responsible for path integration was either MEC II stellate cells or MEC III.
Path integration is calculated using grid cells~\citep{Gil2018}.
Their firing pattern represents metric properties, and the firing pattern of place cells represents the position index information without metric properties~\citep{Buzsaki2013,Fukawa2020}. 
The prospective speed signal is calculated by MEC III or MEC Vb.
Furthermore, in the hippocampus, there is a loop structure wherein the movement signal by path integration in MEC and signal from the LEC are integrated to obtain accurate position information. 
DG and CA3 execute the functions of pattern separation and pattern completion, respectively \citep{Bakker2008}.
Pattern separation is the ability to identify the difference between two perceptual patterns, and pattern completion is the ability to generalize and complement using similar signals from partial observation and noisy environments.

\begin{figure}[!tb]
    \centering
    \includegraphics[width=0.55\textwidth]{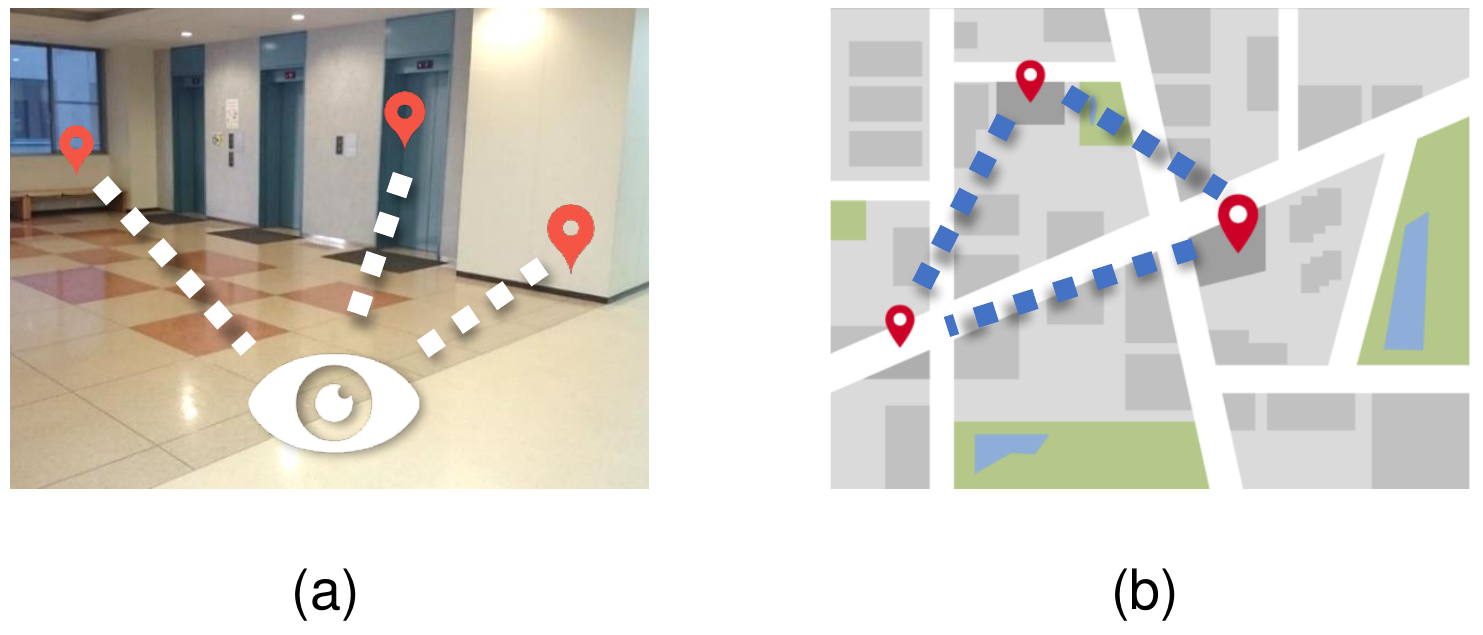}
    \caption{
        Egocentric and allocentric representations of space.
        (a) Egocentric visual information. First-person perspective and self-centered coordinate system.
        (b) Allocentric visual information. Objective perspective and object center coordinate system .
    }
    \label{fig:map_and_view}
\end{figure}

LECs and MECs have been reported to perform different processing functions owing to the difference in the signals sent from PER and POR.
LEC and MEC process proximal and distal landmarks, respectively~\citep{Kuruvilla2020}.
In addition, LEC processes signals from an egocentric perspective, whereas MEC processes signals from an allocentric perspective~\citep{Hargreaves2005,Byrne2007,Deshmukh2011,Bicanski2018a,Alexander2020,Wang2020}.
Figure~\ref{fig:map_and_view} shows the difference between egocentric and allocentric perspectives.
Considering these factors, two types of cognitive maps are generated: survey and route maps~\citep{Shemyakin1962OrientationIS}.
Therefore, LEC and MEC are expected to be responsible for spatial cognition of the route and survey maps, respectively.
These two types of views are used in car navigation systems and online map apps (e.g., Google Street View).

\section{Relevant topics and correspondence between HF and SLAM}
\label{sec:hpf-slam}

We introduce the research on computational models related to spatial cognition methods, including SLAM.
Furthermore, we discuss the association between functions of HF and computational models based on the content covered in \ref{apdx:pgm_slam} and Section~\ref{sec:neuroscience}.
First, we introduce the HF computational models and brain-inspired SLAM in Section~\ref{sec:hpf-slam:models}.
Second, we introduce the models relating to place category formation and investigate their relationship with HF in Section~\ref{sec:spatial_concept}.
Finally, the neural-network models and their associations with PGMs are described in Section~\ref{sec:hpf-slam:neuralnet}.

\subsection{Computational models for HF and brain-inspired SLAM}
\label{sec:hpf-slam:models}

Herein, we describe the HF computational model and the relationship between HF and SLAM.
Several studies have discussed the relationship between SLAM, partially observable Markov decision process graphical models, their probabilistic inferences, and HF circuits and functions~\citep{Penny2013,Madl2018,Gaussier2019,Bermudez-Contreras2020}.
SLAM functions mainly correspond to the loop circuits of the MEC and hippocampus~\citep{Fukawa2020}. 
\citet{Bermudez-Contreras2020} discussed the association with artificial neural networks and reinforcement learning in spatial navigation.
Furthermore, several studies were conducted on computational models of functions relating to the hippocampus from the perspective of computational neuroscience~\citep{Schapiro2017,Banino2018,Kowadlo2019,Scleidorovich2020}.
Several hippocampus-inspired SLAM methods have been proposed~\citep{milford2004ratslam,Tang2018,Yu2019,Zou2020}.
\citet{milford2004ratslam} implemented a biologically inspired mapping system, RatSLAM, which is related to place cells in the hippocampus of a rodent.

Some of the existing SLAM functions can be partially associated with each functional cell in the HF.
Metric representation in the MEC is approximated by the map coordinate system in SLAM.
Grid cells represent spatial metrics to determine coordinate axes~\citep{Hafting2005gridcells}.
Models of grid-cell spatial firing have been proposed in computational neuroscience studies \citep{Zilli2012}.
In contrast, the map representation in numerous SLAM models is either a 2D or 3D Cartesian coordinate system.
Particularly, the occupancy-grid map of SLAM can be regarded as a combined representation of grid and border cells.
Additionally, in the self-localization model designed for engineering~\citep{ishibushi2015statistical}, Gaussian distributions are arranged at equal intervals, resulting in a representation resembling grid cells.
The state variables in SLAM that represent posture are defined by position coordinates and orientation.
Therefore, posture is associated with the firing of grid and head-direction cells.

In landmark-based SLAM, which is a time-series state-space model, the landmark position is estimated using a Kalman filter~\citep{montemerlo2002fastslam}.
This corresponds to the MEC function for processing distal landmark signals obtained from POR~\citep{Kuruvilla2020}.
In general SLAM, it is assumed that the landmarks are static.
Expanding this to dynamic objects is expected to enable the estimation of the position of the movement of other individuals, for example, as a dynamic event that is part of the external context in the environment.
This type of expanded SLAM model may be a candidate model for a research report on the activity pattern of place cells in other rats~\citep{Danjo2018}.

In the brain, self-localization is performed by theta-phase precession~\citep{Hafting2008}.
Phase precession is explicitly modeled as a spatial cell model in \citet{Zou2020}.
However, in the SLAM studies, there are a few models that explicitly implement phase precession, even in brain-inspired models.
Hence, it is implemented by different processes that are functionally similar in engineering.
In this study, we propose a discrete-event queue, as shown in Section~\ref{sec:hpf-pgm:queue}.

Episodic memory is an important function both in the hippocampus and robotics.
In the computational neuroscience field, models of the hippocampus in episodic memory have been proposed~\citep{Mcnaughton1987,Treves1994,ME1997}.
Several methods incorporating episodic memory have been proposed for robotics~\citep{Tang2017,Ueda2018,ref:furuta2018everyday,Zou2020}.
\citet{Ueda2018} proposed a brain-inspired method (i.e., a particle filter on episode) for agent decision-making.
An episode that replays in CA3 may be modeled with a generative process.
There are various ways to express episodic memory; however, robots can retain temporal transitions with observed events and refer to them subsequently.

\subsection{Semantic mapping and spatial concept formation in robotics}
\label{sec:spatial_concept}

In mobile robots, it is essential to appropriately generalize and form place categories while dealing with the uncertainty of observations.
Hence, a semantic mapping approach, including semantics of places and objects, has been actively developed~\citep{kostavelis2015semantic,Garg2020}.
To address these issues, PGMs for spatial concept formations have been constructed \citep{isobe2017learning,ataniguchi_IROS2017,hagiwara2018hierarchical,ataniguchi2020spcoslam2,Katsumata2020SpCoMapGAN}.
\citet{ataniguchi_IROS2017,ataniguchi2020spcoslam2} realized a PGM for online spatial concept acquisition with simultaneous localization and map-ping (SpCoSLAM), which conducts place categorization and mapping through unsupervised online learning from multimodal observations.
Visual information is used as landscape features reminiscent of spatial view cells~\citep{Rolls2013}.
HF is also centrally involved in the formation of place categories, semantic memory, and understanding of the meaning of places by integrating signals from each sensory organ~\citep{Buzsaki2013}. 
We consider the aforementioned models as candidates for a cognitive module with functions similar to those of the HF.

SpCoSLAM is a model that arguably imitates some functions of the hippocampus and cerebral cortex.
From the viewpoint of computational efficiency and estimation accuracy, \citet{ataniguchi2020spcoslam2} proposed an inference algorithm that sequentially re-estimates some recent events and accordingly updates global parameters in older observations. 
Assuming that the training data (i.e., the event-based robotic experience) represent episodic memory and spatial concepts represent semantic memory, their algorithms can sequentially extract concepts from short-term episodic memory to form a semantic memory.
\citet{isobe2017learning} proposed a model for place categorization using self-position and object recognition results.
This model shows that the categorization accuracy is higher when weighing is performed while considering only the objects that are close to the robot, rather than using all objects that are evident from the robot's viewpoint for place categorization. 
This result is also consistent with the neuroscientific findings of proximal landmarks~\citep{Kuruvilla2020}. 
Furthermore, the hierarchical multimodal latent Dirichlet allocation~\citep{Ando2013hMLDA} provides a categorical representation of hierarchical locations~\citep{hagiwara2018hierarchical}.
The multi-layered k-means was adopted to extract the hierarchical positional features of a space.
Arguably, this corresponds to the hierarchical representation of grid cells in an MEC~\citep{Hafting2005gridcells} and is considered a valid model based on neuroscientific findings. 
Although the aforementioned algorithms and models were not originally inspired by biology or neuroscience, such research is highly suggestive.

\subsection{Neural-network models}
\label{sec:hpf-slam:neuralnet}

Deep neural-network models, such as world models~\citep{Ha2018}, represent various types of information in the latent state space.
In ordinary SLAM, the functional shapes of the distributions (i.e., global parameters) for motion models, which represent state transitions and the measurement models that match observations to a map, are designed and set by humans.
Therefore, SLAM problem means the estimation of local latent parameters.
In contrast, world models do not have explicit models embedded with prior knowledge.
Deep neural-network models learn elements corresponding to global and local parameters in PGMs simultaneously.
The estimation of motion and measurement models, including map representations on neural networks, is related to predictive coding.

As computational models for HF, grid cells and their similar spatial representations have been reproduced by deep neural-networks, including long short-term memory (LSTM) or multi-layered recurrent neural networks (RNNs)~\citep{noguchi2018navigation,Grid2018,Banino2018}.
A vector representation similar to that of grid cells is acquired as an internal state by LSTMs, which predicts the current posture from the past velocity and angular velocity with dropouts~\citep{Banino2018}.
In contrast, from an engineering point of view, models for place representation and spatial concepts using deep neural-networks have been proposed.
The room space is learned from the visual-motor experience using two sub-networks comprising a deep auto-encoder and an LSTM~\citep{yamada2017learning}.
By integrating the spatial concept formation model introduced in Section~\ref{sec:spatial_concept} with generative adversarial networks, \citet{Katsumata2020SpCoMapGAN} transferred global spatial knowledge from multiple environments to a new environment.
These approaches have some suggestive elements that can be interpreted as an HF model.

With the advent of Bayesian deep learning and deep PGMs, it has become possible to discuss neural networks within the framework of PGMs.
In particular, by implementing it in the framework of deep PGMs, it is possible to naturally incorporate hierarchical RNNs into a generative model, such as variational auto-encoder and generative adversarial networks.
For example, the predictive-coding-inspired variational RNN~\citep{Ahmadi2019} is not a model for the hippocampus or navigation task, but a hierarchical RNN-based deep PGM.
Deep PGMs are achieved by amortized variational inference, a type of variational inference that introduces functions to transform observation data into parameters of the approximate posterior distribution.
Therefore, it is possible to formulate arbitrary variable transformations and nonlinear functions as probability distributions.
The framework for maximizing the evidence lower bound by variational inference is equivalent to free energy minimization.
Therefore, if the HF models are described by PGMs, they can be naturally connected to the free energy principle~\citep{friston2019} and world models~\citep{Ha2018}.

\section{HF-PGM: Hippocampal formation-inspired probabilistic generative models}
\label{sec:hpf-pgm}

We constructed a graphical model of HF and its surrounding regions: HF-PGM.
We describe the HF-PGM in association with the HF by interpreting and modeling it using a graphical model.
Section~\ref{sec:hpf-pgm:roi} describes the region of interest, top-level function, and input/output signals.
Section~\ref{sec:hpf-pgm:executing_gipa} describes the execution for allocation of generative or inference process in connections.
Section~\ref{sec:hpf-pgm:model} describes the time-series representation, encoder-decoder representation, and representation associated with BIF in the proposed HF-PGM.
Section~\ref{sec:hpf-pgm:step3} discusses the consistency of the model with scientific knowledge.

\subsection{Region of interest, top-level function, and input/output}
\label{sec:hpf-pgm:roi}

The region of interest targeted in this study to realize the proposed model includes LEC, MEC, CA1, CA3, DG, Sb, and ParaSb.
Additionally, adjacent areas of the region of interest (i.e., areas where the signal can be obtained or a signal is sent) are PreSb, POR, PER, and RSC.
The connection relationship of each region is based on Fig.~\ref{fig:hpf_circuit}.
\citet{Fukawa2020} mainly focused on the MEC and hippocampal circuits.
In addition, LEC is considered in this study.
Hypotheses on LEC functions were derived from several papers and used for model construction. 
This study assumes that the model works when the agent is walking in an awake state.
The top-level function for the assumed region of interest integrates allocentric and egocentric information once and outputs each prediction.
The following activities describe the processing inside the model used to realize the top-level function:
\\
{\textbf{(i) Self-localization (path integration and observational correction)}:
Self-localization is performed by path integration and observation-based prediction correction (as explained in \ref{apdx:pgm_slam:SLAM} and Section~\ref{sec:neuroscience:functions}).
The difference from conventional SLAM is that the prediction is corrected using integrated information from the LEC-variables, and the prediction of the future times is output to other regions. 
For details, refer to Section~\ref{sec:hpf-pgm:model:pgm-time}.
\\
\textbf{(ii) Place categorization by integrating allocentric and egocentric information}:
By integrating the allocentric visual information processed by the MEC and the egocentric visual information processed by the LEC, a place category representing semantic memory about a place is formed~\citep{Buzsaki2013}.
As a PGM, it can be modeled as a multimodal categorization.
For details, refer to ``integration of information'' in Section~\ref{sec:hpf-pgm:model:pgm-bif}.}

The inputs to the model include variables in POR, PER, and RSC.
The internal representations at these parts are treated as observed variables in PGM.
The outputs of the model are the predicted values in POR and PER at the next time-step.
In PGM, the latent variables and parameters of the conditional distribution are obtained as the estimated values.
The definition of each corresponding variable is described in Table~\ref{tab:element_of_graphical_model}.

\subsection{Executing generation-inference process allocation}
\label{sec:hpf-pgm:executing_gipa}

We constructed HF-PGM to be consistent with SLAM's PGM based on the generation-inference process allocation procedure (see Section~\ref{sec:hpf-pgm:gipa}). 
The connection between POR and MEC II superficial on the BIF (Fig.~\ref{fig:hpf_circuit}) can be regarded as a feedforward pathway.
An inference process can be allocated to this connection in the PGM.
In contrast, the connection between MEC V and POR on the BIF can be regarded as a feedback pathway.
A generative process can be allocated to this connection in the PGM.
The connections between LEC and PER are assigned in the same way.

There exist limitations to performing generation-inference process allocation for the inside of the hippocampus while considering only the connectivity with the neocortex.
Therefore, the engineering formulation of the SLAM modeled as PGM is used as a reference.
Partially following the generative process in the PGM of SLAM (see \ref{apdx:pgm_slam:SLAM}), the links related to the state transition, motion models $p(x_{t} \mid x_{t-1}, u_{t})$, were assigned as the generative processes.

\subsection{Models and operating principle}
\label{sec:hpf-pgm:model}

\begin{figure}[!tb]
    \centering
    \includegraphics[width=1.0\textwidth]{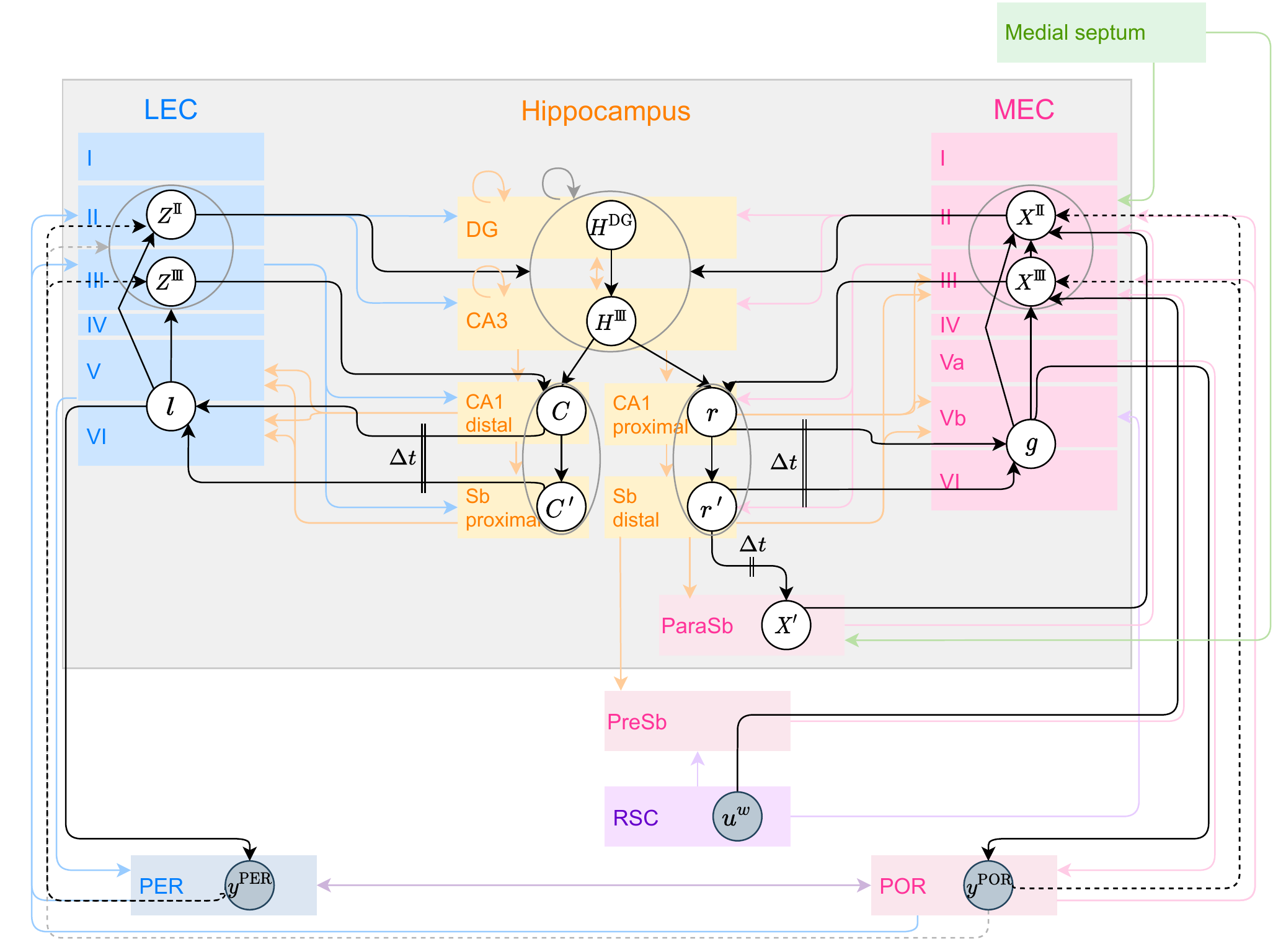}
    \caption{
        Graphical model representation of HF-PGM {as the final form}. 
        It is drawn on the BIF, representing the circuit diagram of the HF shown in Fig.~\ref{fig:hpf_circuit}. 
        The area surrounded by a gray frame is the region of interest targeted in this study. 
        Gray nodes represent observed variables, whereas white nodes represent unobserved latent variables.
        Black arrows represent the generative process, and dotted arrows represent the inference process.
        The flat arrow with $\Delta t $ indicates the generation of the variable in the next time step.
        Nodes surrounded by gray circles are assumed to be functionally similar and may be treated as the same variable. 
    }
    \label{fig:hpf_pgm}
\end{figure}

In this section, we describe the model structure and function of the proposed HF-PGM.
We associate variables with each region of the HF.
Some anatomical connections are omitted in the PGM for engineering feasibility.
Descriptions of global parameters are omitted from the graphical model.
In this study, a specific shape is not particularly limited in probability distributions.
Only random variables, their functional definitions, and dependencies between the random variables are assumed.
Here, the node does not necessarily have to be a random variable.
Signals from outside the HF-PGM module are treated as observed variables, and it is assumed that the observation of each input is converted from raw data into individual high-level features through each processing module; additionally, it enters the HF-PGM.
Models may be considered from raw data that can be observed by robots; however, because we intend focus on the top-level function, for the time being, we consider them as extracted features.

The {final form of }HF-PGM is shown in Fig.~\ref{fig:hpf_pgm}.
HF-PGM was expressed using inference and generative processes based on the generation-inference process allocation. 
We describe the variables of HF-PGM in Table~\ref{tab:element_of_graphical_model}.
Next, two types of descriptions in the table are described.
Physiological findings include a description of a physiological phenomenon observed at a specific site on the BIF or the function inferred from it.
Because there are many uncertain elements in physiological findings related to LECs, the description is omitted here.
The function of the components is a description of the computational functions assumed for each component included in the HCD. 
As a detail of the aforementioned contents with references, we released a pre-screening version of the BRA data\footnote{Hippocampal formation BRA data (pre-screening version): \url{https://docs.google.com/spreadsheets/d/1xf5tIj2qzHh9a52a2p9K5b8ggra825bvBdXKhghF2W4/edit\#gid=0}} \addspan{as supplementary material (\texttt{hf\_bra.xlsx})}, which describes the navigation functions of the HF.

Sections~\ref{sec:hpf-pgm:model:pgm-time} -- \ref{sec:hpf-pgm:model:pgm-bif} explain the flow of changes from the general SLAM models in a step by step manner, as shown in Fig.~\ref{fig:hpf_pgm}. 
This implies that the functions of SLAM's PGM can be decomposed and associated based on the anatomical structure of HF. 
Meanwhile, the functions of the HF-PGM can include the functions of the conventional SLAM (discussed in Section \ref{sec:hpf-slam:models}) and the spatial concept formation models (introduced in Section \ref{sec:spatial_concept}).
The operating principle, as an internal process for realizing top-level function in the HF-PGM, is shown below. 
These also serve as sub-functions subordinate to the top-level function.

{\scriptsize
\begin{tabularx}{\linewidth}{p{11mm}X||p{9mm}p{45mm}}
    \caption{
        Description of variables in the HF-PGM.
        Grey backgrounds are associated with PGM for SLAM, as shown in Fig.~\ref{fig:slam_gm}. 
        The corresponding variables are listed together in symbol.
    }
    \\ \hline
    \textbf{Region} & \textbf{Physiological findings} & \textbf{Symbol} & \textbf{Function of components on HCD} \\
    \hline
    CA1\, distal   & Non-spatial semantic memory & $C$             & Place category (internal representation of visual spatial information) \\ 
    CA1 proximal & Place cells & $r$              & Position distribution (cluster information regarding positions) \\
    Sb\, proximal  & State at CA1 distal with time delay   & $C^{\prime}$   & Place category at the previous time \\
    Sb distal    & State at CA1 proximal with time delay & $r^{\prime}$   & Position distribution at the previous time \\
    CA3          & Pattern completion, information integration & $H^{\III}$      & Integrated semantic memory and episodic memory of information from $X$ and $Z$ \\
    DG           & Pattern separation                     & $H^{\rm DG}$  & Integrated semantic memory \\
    \rowcolor[gray]{0.9}
    MEC \II      & Grid cells, Border cells, path integration & $X^{\II}$ ($\{x_{t}\}$)      &  Self-position information, predictive distribution \\
    \rowcolor[gray]{0.9}
    MEC \III     & Grid cells, head-direction cells, border cells & $X^{\III}$ ($\{x_{t}\}$)      & Self-posture information (position and orientation), observation likelihood \\ 
    MEC Va,Vb,VI & Prospective speed calculation, feedback to POR & $g$             & Predictive representation of $X$ (Prediction at future time regarding movement/speed amount or posture) \\
    LEC \II      & --- & $Z^{\II}$       & Abstraction of information from $y^{\rm{PER}}$ (transmission of prediction, generation of prediction signal) \\
    LEC \III     & --- & $Z^{\III}$      & Abstraction of information from $y^{\rm{PER}}$ (Observation transmission) \\
    LEC V,VI     & Feedback to PER & $l$ & Predictive representation of $Z$ (Prediction at future time from the difference between $C^{\prime}$ and $C$) \\
    \rowcolor[gray]{0.9}
    ParaSb       & --- & $X^{\prime}$ ($\{x_{t-1}\}$)    & Self-posture information \\
    \rowcolor[gray]{0.9}
    POR          & --- & $y^{\rm{POR}}$ ($\{y_{t}\}$) & Allocentric visual information (distal distance/landmarks, absolute object positions) \\
    PER          & --- & $y^{\rm{PER}}$  & Egocentric visual information (proximal distance/landmarks, relative object positions, object category, landscape information) \\ 
    \rowcolor[gray]{0.9}
    RSC          & Head direction signal & $u^{w}$ ($\{u_{t}\}$)        & Rotational speed movement \\ 
    \hline
    \label{tab:element_of_graphical_model}
\end{tabularx}
}

\subsubsection{PGM representation according to hippocampus and MEC structures}
\label{sec:hpf-pgm:model:pgm-time}

\begin{figure}[!tb]
    \centering
    \includegraphics[width=0.7\textwidth]{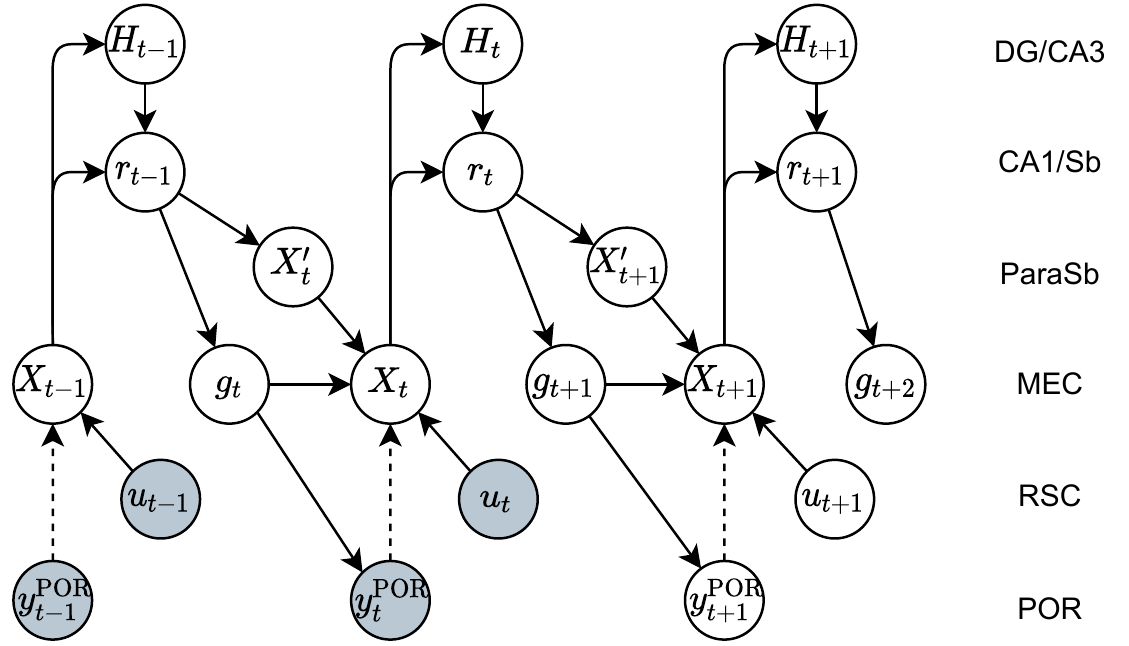}
    \caption{
        Time-series version of graphical model representation of only the MEC side in HF-PGM. 
        The generation and inference processes are drawn simultaneously.
        The subscript representing time reflects the time of the outside world, in which the observation is obtained.
        Notably, it does not represent the time of internal processing.
        Observations up to the current time, $ t $, are obtained. 
    }
    \label{fig:hpf_pgm_time}
\end{figure}

The graphical model is shown in Fig.~\ref{fig:hpf_pgm_time}, which is a variant of the PGM of SLAM (see Fig.~\ref{fig:slam_gm}) and is mapped with reference to the hippocampal and MEC loop structures.
This model is an extension of a partially observable Markov decision process.
In this model representation, we omit the connections on the LEC side. However, the signal obtained from the LEC side was originally integrated via $H_{t}$, which corresponds to the integrated higher-level internal representation.
Corresponding to this graphical model of the neural circuits in the hippocampus and MEC, we obtain the following:
CA1/Sb ($r_{t-1}$)
$\rightarrow$
ParaSb ($X^{\prime}_{t}$)
\&
MEC Vb, VI ($g_{t}$)
+
POR/RSC ($y_{t}^{\text{POR}}, u_{t}$)
$\rightarrow$
MEC \II (${X}_{t}$)
$\rightarrow$
DG/CA3 ($H_{t}$)
\&
CA1/Sb ($r_{t}$)
$\rightarrow$
MEC Vb, VI ($g _{t+1}$)
$\rightarrow$
POR ($\hat{y}_{t+1}^{\text{POR}}$).
Some are defined as the same variable, assuming that they have similar functions; however, the regions are different. 

HF-PGM differs from the traditional PGM of SLAM when generating predictions of observation for the next cycle.
SLAM directly estimates the current self-posture, $x_{t}$, from the previous self-posture, $x_{t-1}$.
Furthermore, the generative process of SLAM is the arrow from $x_{t}$ to $y_{t}$ at time $t$.
In contrast, HF-PGM generates the self-posture, $X_{t+1}$, from $X_{t}$ via variables, such as $H_{t}$, $r_{t}$, $X^{\prime}_{t+1}$, and $g_{t+1}$.

The higher-level representation of place is denoted as $r_{t}$.
$g_{t+1}$ is responsible for the prediction of the next time.
It is assumed that $r_{t}$ comprises differential information about the time delay between CA1 and Sb.
Hence, the prediction, $g_{t+1}$, is generated from $r_{t}$, and the time-differential information is used.
This prediction by the difference calculation is not performed in conventional SLAM.

Additionally, $r_{t-1}$ at the previous time, $t-1$, generates the self-posture, $X^{\prime}_{t}$.
$X^{\prime}_{t}$ is responsible for conveying the position information of the previous time.
The self-position, $X_{t}$, is predicted from $X^{\prime}_{t}$ and the movement amount, $u_{t}$.
Subsequently, the self-position, $X_{t}$, is corrected by the predicted value, $g_{t}$, and the observation, $y_{t}^{\text{POR}}$.

It is assumed that the predicted value, $g_{t}$, is determined by reducing the error so that $X_{t}$ can be generated consistently with other variables.
It is also possible to minimize the prediction error of $\hat{y}_{t}^{\rm POR}$, which is generated from the prediction, $g_{t}$, and the actual observation, $y_{t}^{\rm POR}$: predictive coding~\citep{Rao1999}.

Related to the theta-phase precession phenomena in the HF circuit, signals that circulate in the looping circuit in HF require units of discrete-event queues in a current state, which includes both near-past and near-future predictions (see also Section~\ref{sec:hpf-pgm:queue}).
Additionally, given the nature of theta-phase precession, variables that form a large loop may not have self-transition~\citep{Butler2018}.
Therefore, it is assumed that $H_{t}$, $r_{t}$, and $X_{t}$ do not have a direct time self-transition despite having a direct neural connection.

\subsubsection{Integrated with LEC}
\label{sec:hpf-pgm:model:pgm-mae}

\begin{figure}[!tb]
    \centering
    \includegraphics[width=0.8\textwidth]{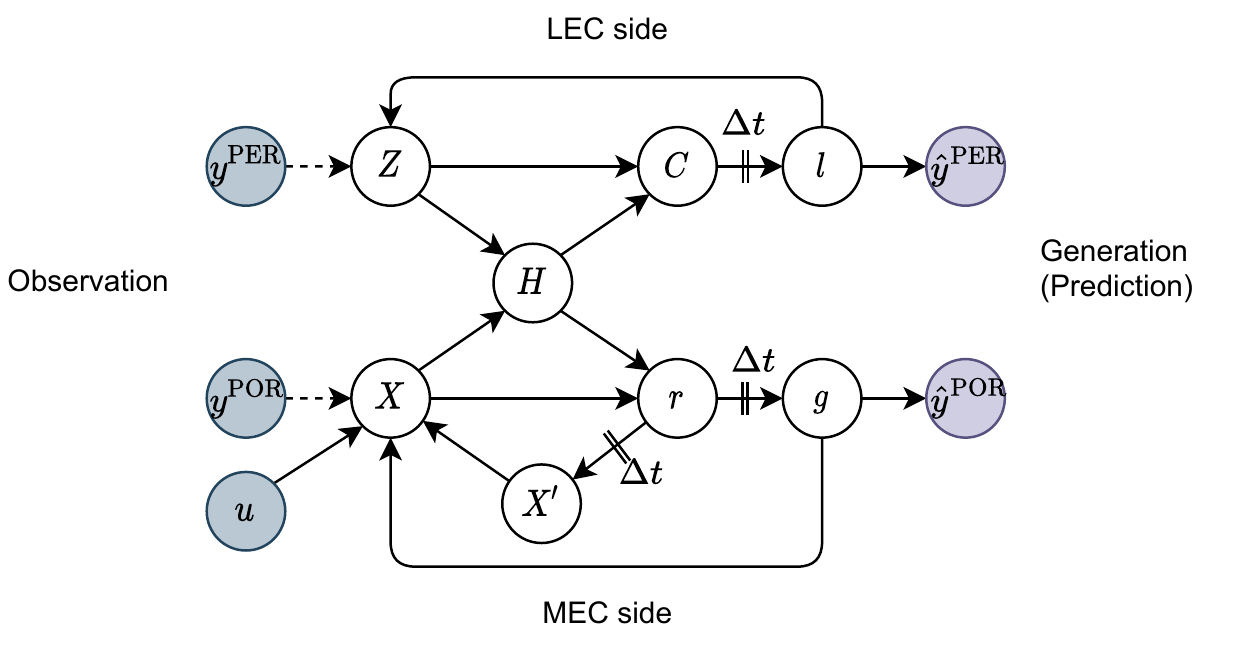}
    \caption{
        Encoder--decoder version of graphical model representation in HF-PGM. 
    }
    \label{fig:hpf_pgm_MAE}
\end{figure}

Figure~\ref{fig:hpf_pgm_MAE} is a time-omitted representation of the graphical model in Fig.~\ref{fig:hpf_pgm_time} with additional connections on the LEC side.
The projection from LEC V, VI $l$ to LEC II, III $Z$ assumes a connection structure similar to MEC.
This graphical model representation clearly shows the result of generation-inference process allocation.
Because the HF-PGM in Fig.~\ref{fig:hpf_pgm_MAE} is compressed in time, a circulation occurs in the generative process.
Therefore, the notation for the next time generation process introduced in Fig.~\ref{fig:slam_gm} is used.
Notably, there exists an arbitrariness of position with which the time progress can be allocated in the PGM loop.

This model has an encoder--decoder structure and can be seen as an extended form of variational encoders--decoders~\citep{Bahuleyan2018} with two modalities and a condition.
Because the model structure is similar to encoder--decoder models, the input and output have the same variables with the next time generation process for the inference and generation sides.
The signals (${y}^{\text{PER}}$ and ${y}^{\text{POR}}$) from PER and POR-parts are processed into $Z$ and $X$ by LEC and MEC, respectively, integrated into $H$ by DC/CA3-part and divided into $C$ and $r$. 
Subsequently, $\hat{y}^{\text{PER}}$ and $\hat{y}^{\text{POR}}$ are returned from the predictive information, $l$ and $g$.
During training, the loss function can be designed to match the input to the generated output.
The structures of $Z$ to $C$ and $X$ to $r$ are similar to those of the skip connection (i.e., the contracting path) and the U-net~\citep{Ronneberger2015}.
Furthermore, a loop structure has a recurrent time delay on the MEC side, which is consistent with the anatomy of the MEC and hippocampus.
It is suggested that this loop structure is crucial to self-localization~\citep{Fukawa2020}.

\subsubsection{HF-PGM associated with BIF}
\label{sec:hpf-pgm:model:pgm-bif}

Finally, corresponding to the BIF of the HF, Fig.~\ref{fig:hpf_pgm} is obtained.
The following is a detailed description of each part of the HF-PGM based on Fig.~\ref{fig:hpf_pgm}.

\textbf{Input and output}: 
The amount of rotational movement, $u_{t}^{w}$, is transmitted from RSC to MEC \III ($X^{\III}$) via PreSb.
The translational speed is assumed to be calculated inside the MEC, where the speed cells exist~\citep{Kropff2015, Hinman2016}, because the corresponding region cannot be clearly identified.
Hence, the amount of movement, $u_{t}$, is obtained by integrating the difference information sent from $r$ and $r^{\prime}$ with $u_{t}^{w}$.
In engineering, it is possible to calculate the speed from the difference information. For example, there are methods such as optical flow.
A discussion related to the aforementioned sentences is provided in \ref{apdx:future:speed}.
Additionally, this time-difference information, as mentioned in Section~\ref{sec:hpf-pgm:model:pgm-time}, is useful for predicting the internal state and input signal at the next time\footnote{As the variables in CA3 and Sb are separated in Fig.~\ref{fig:hpf_pgm}, it is considered that C and r hold time-difference information as an internal representation.
Because $r^{\prime}$ and $C^{\prime}$ are said to represent the state having a time delay of $r$ and $C$ in BIF, further studies are required to assess the assumptions on the separability and generation of these variables.}.
POR $y^{\rm POR}$ mainly deals with distal landmarks and distance signals~\citep{Kuruvilla2020}.
Distant information is useful for self-localization because it can be obtained more robustly than proximal information while moving.
PER $y^{\rm PER}$ mainly deals with proximal landmarks and distance signals.
This signal is useful to avoid proximal obstacles.
Additionally, proximal objects may be useful in forming a place category for the current location~\citep{isobe2017learning}.
Landscape information can also be used to roughly identify the current location~\citep{Rolls2013}.
These are similar to the treatments of observations in spatial concept formation, as described in Section~\ref{sec:spatial_concept}.

\textbf{Role of LEC/MEC}: 
Generally, in the neocortex, layer \III is thought to be responsible for the transmission of observations, and layer \II for predictions~\citep{Yamakawa2020}.
Given that the LEC/MEC is also part of the neocortex, its role is likely to be preserved.
LEC/MEC \III receives observation signals from the POR and PER and projects them mainly to CA1 and other areas.
LEC/MEC \II generates predictive signals and projects them to DG and CA3.
Hence, it is considered that LEC/MEC \II mainly calculates the predictive distribution, and LEC/MEC \III calculates the observation likelihood.
Because the MEC has a coordinate system with grid cells, $X$ represents the robot's posture in the environment.
MEC \II, \III receives the observation, $y^{\rm POR}$, from the POR and the rotational movement, $u_{t}^{w}$, from the RSC.
Egocentric visual information $y^{\rm PER}$ (e.g., proximal object signal~\citep{Kuruvilla2020}) is sent to $Z^{\II}$, $Z^{\III}$ in LEC.
The variable $l$ of LEC V, VI is expected to be latent variables that serve as intermediates to send a generative signal to the PER.

\textbf{Integration of information}: 
Allocentric and egocentric information is integrated into the hippocampus.
Hence, DG $H^{\rm DG}$ and CA3 $H^{\rm III}$ form an abstract internal representation of a place that integrates information from the visual information, $Z$, in LEC and the positional information, $X$, in MEC.
We believe that this internal representation corresponds to the spatial concepts.
Furthermore, DG and CA3 are said to have functions of pattern separation and completion, respectively ~\citep{Bakker2008}.
{These functions are crucial in the integration of multimodal information.}
From the PGM perspective, pattern separation may be modeled by parameters that determine the Bayesian prior distribution~\citep{Sanders2020}.
Specifically, the concentration parameter in Dirichlet process clustering is involved in the automatic determination of the number of clusters~\citep{Neal2000a}.
{In short, allocentric and egocentric information are expected to form a cluster within a unified latent space.}
Pattern completion can be viewed as the process of regenerating information from DG as defect completion by resampling from a probability distribution.
Additionally, error correction of self-position is believed to occur in the hippocampus~\citep{Fukawa2020}.
By this process, simultaneously integrated information $H^{\rm DG}$ and $H^{\III}$ can be used.
We assume that place-category formation or information processing occurs in the CA1 distal region.
Further, we assume that location-dependent category formation corresponding to place cells occurs in the CA1 proximal region.

\subsection{Consistency of model with scientific knowledge}
\label{sec:hpf-pgm:step3}

This section discusses the consistency of the model using scientific knowledge{ and how the outcome can be tested}.
We believe that the HF-PGM is highly feasible because it is consistent with the anatomical findings of HF, although there may be more detailed variables and dependencies.
The main reason is that the original PGM of SLAM already exists.
We also follow the path integration of MEC and hippocampus, as discussed in \citet{Fukawa2020}.
Furthermore, the agreement with BIF and engineering operating principles is described in Sections~\ref{sec:hpf-pgm:model}.
Consequently, we successfully map the HF as a PGM by introducing the generation-inference process allocation.
In summary, HF-PGM is highly consistent with the brain structure of HF.

The effectiveness of HF-PGM can be verified by solving tasks that can be achieved by integrating LEC (egocentric signal) and MEC (allocentric signal). For example, in situations where localization is difficult in SLAM, we can investigate whether the place category information can distinguish the position. In addition, HF-PGM can be realized as a concrete model by integrating the neural network-based world model and SLAM. Then, further experiments can be conducted to determine whether the latent space representation of the world model, which tends to be unstable in learning, can be complemented by the geometric information of SLAM.
\addspan{
We have also provided a pseudo-source code in the form of Neuro-SERKET architecture~\citep{Taniguchi2020neuro} for HF-PGM as supplementary material (\texttt{HF-PGM\_serket.py}).
}

The implementation of HF-PGM on a robotic platform is a hot research topic for future studies.
Hence, we plan to explore the detailed structure of each model element in subsequent studies.
The HF-PGM proposed in this study is a specimen, and the actual implementation may require more concrete discovery in terms of engineering. 
Therefore, the following issues must be mitigated, including 
(i) selection of a type for each probability distribution during the generative process,
(ii) when performing amortized variational inference, selection of the function shape of the inference model and that of the architecture of the neural networks, 
and
(iii) ensuring a real-time algorithm that includes the learning of global parameters.
The aforementioned issues can be solved by model selection and architecture search/optimization in a framework similar to neural architecture search.

\section{Abstraction as discrete-event queue} 
\label{sec:hpf-pgm:queue}

We provide the interpretation of the \textit{phase precession queue assumption}, which is one of the functionalities of the HF as an estimation of the probability distribution in PGM.
As explained in Section~\ref{sec:hpf-pgm:model:pgm-time}, a discrete-event queue based on this assumption was constructed to explain the parts that cannot be expressed by the graphical model structure of the PGM alone.
The phase precession queue assumption plays an important role in the development of an HCD from BIF. Thus, its relationship with the PGM framework is indirect. 

\begin{figure}[!tb]
    \centering
    \includegraphics[width=1.00\textwidth]{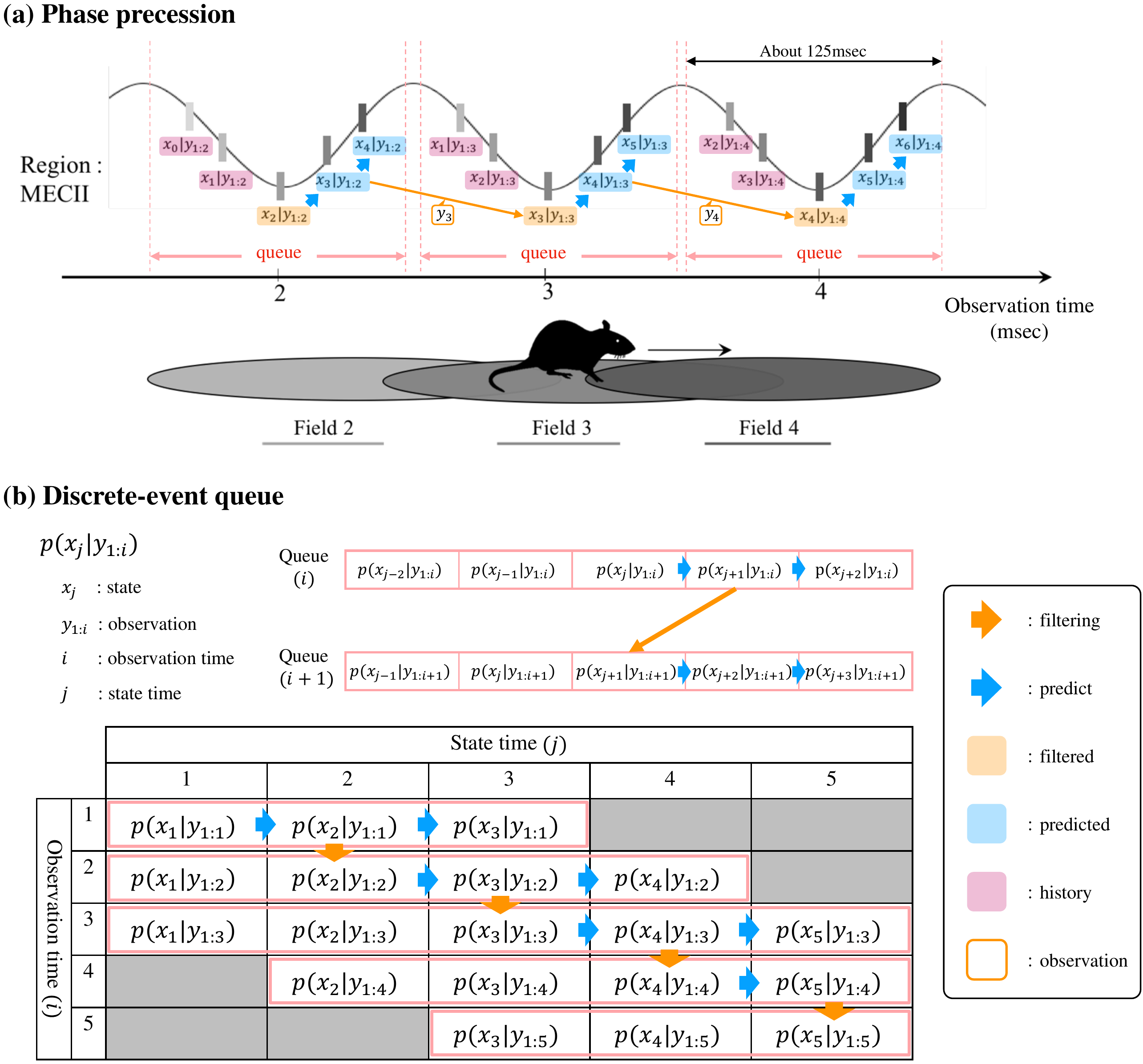}
    \caption{
        (a) Phase precession queue assumptions and (b) discrete-event queue processing.
        (a) shows the information representation in the theta-phase precession of MEC \II when the rat moves from left to right.
        $x_{3} | y_{1:3}$ represents an estimate of the state, given the observations up to time $t = 3$.
        Fields 2--4 at the bottom of the rat represent the spatial regions where the three grid cells fire in place fields.
        The horizontal axis of the bottom table in (b) represents the time of the estimated state, and the vertical axis represents the time of observation.
        The red box represents the queue.
        The queue has the estimated/predicted states five times from $t-2$ to $t + 2$ at the current time, $ t $.
        The vertical axis of the table in (b), which is the observation time, corresponds to the horizontal axis in (a), which is the actual time. 
    }
    \label{fig:hpf_queue}
\end{figure}

In the HF, the theta-phase precession is known to process the experience by discretizing it and compressing it within a time step, as shown in Figure~\ref{fig:hpf_queue}~(a).
Herein, stimuli from the external world are sampled at the period of theta waves (8--13 Hz), and it is believed that the present, past, and future events are encoded in phase \citep{Terada2017}.
By abstracting the information within one phase as a queue, as shown in this subsection, the calculation of the queue can be interpreted as filtering the current state, smoothing the past, and predicting the future.

\subsection{Introducing discrete-event queue}
\label{sec:hpf-pgm:queue:intro}

To comprehensibly model the compressed time process handled by the theta-phase precession, the time granularity of the entire model must be detailed.
To avoid too much complexity, we introduce the phase precession queue assumption, which states that the signal held by the theta-phase precession can be regarded as a time queue containing the past, present, and future events at the current time.
The assumption is as follows:

\begin{quote}
\textbf{Phase precession queue assumption}:   
The observed state compressed within one cycle of theta waves circulating on a pentasynaptic loop circuit can be regarded as a discrete-event queue representing the state sampled at discrete time intervals. Here, the pentasynaptic loop circuit is formed in the hippocampus and MEC by the projection sequence MEC\,II-DG-CA3-CA1-Sb-ParaSb-MEC\,II.
\end{quote}

There are several reasons why the aforementioned assumption would be considered reasonable.
As a neuroscientific finding, HF is expected to be modeled as a discrete-event queue with a finite buffer capacity according to the analysis of large-scale network communication in macaque monkeys~\citep{Misic2014}.
Further, in intelligent systems that deal with a world with hidden Markov properties, it is useful to have a discrete-event queue capability to retain the entire observed signal for a short time.
From an engineering perspective, a discrete-event queue can be easily realized as a memory array for the number of time steps to be stored. Here, the memory elements on the same array are considered to have the same meaning, but at different times. However, the neural circuitry of the brain has a restriction that it cannot have more than one representation with the same meaning. Therefore, it seems that we have no choice but to use a method like the theta-phase precession, which compresses information in the time dimension on the same neuron.
The fact that PraSb and MEC II on the pentasynaptic loop circuit receives direct projections from the medial septum, which generates theta rhythms, is consistent with the assumption that this circuit is involved in the theta-phase precession.

\subsection{Processing for discrete-event queue}
\label{sec:hpf-pgm:queue:formulation}

From the aforementioned discussion, the information held in the phase precession can be interpreted as a discrete-event queue.
If this is the case, the process shown in Fig.~\ref{fig:hpf_queue} (b) is performed. 
Theoretically, the discrete-event queue can be regarded as a sequential estimation problem for the joint posterior distribution in multiple states.
The variables are the same as those of the PGM of SLAM, as shown in Fig.~\ref{fig:slam_gm}.
Table~\ref{tab:element_of_graphical_model} lists the correspondence of variables with HF-PGM.
Notably, $X_{t}$ in the HF-PGM is a variable with an internal representation equivalent to $x_{t-2:t+2}$.
Figure~\ref{fig:hpf_queue} (b) shows a table presenting a simplified notation for the discrete-event queue.
Each element in the table is a conditional marginal probability distribution in the state, $x_{j}$, at time $j$ under the condition of observations up to time $i$.
Here, the control value, $u_{t}$, and the integrated information from the LEC are omitted.
The probability distribution of the discrete-event queue is shown in Equation (\ref{eq:queue_prob}). 
\begin{eqnarray}
\text{Queue}(t) &=& p(x_{t-2:t+2} \mid y_{1:t}), \qquad j \in \{ i-2 \leqq i \leqq i+2 \}, \quad i=t, \nonumber 
\\ &=&  \eta \underbrace{p(x_{t+2} \mid x_{t+1})}_{\text{Prediction}} \underbrace{p(y_{t} \mid x_{t})}_{\text{Filtering}} \underbrace{p(x_{t-2:t+1} \mid y_{1:t-1})}_{\int \text{Queue}(t-1) dx_{t-3}}, \quad t \geqq 3.
\label{eq:queue_prob}
\end{eqnarray}
Here, we assume that $\text{Queue}(t-1)$ is calculated at the previous time, $t-1$.
$i$ and $ j$ in the formula correspond to those in Fig.~\ref{fig:hpf_queue}.
${\int \text{Queue}(t-1) dx_{t-3}}$ denotes the operation of the integrated-out (i.e., marginalizing)~\citep{murphy2012machine} of $x_{t-3}$ from $\text{Queue}(t-1)$.
As shown in Eq.~(\ref{eq:queue_prob}), $\text{Queue}(t)$ is a recurrence formula expressed by $\text{Queue}(t-1)$, and a sequential calculation similar to the Bayes filter is possible.

The discrete-event queue can be interpreted as an algorithm that combines the filtering with the smoother and the prediction~\citep{kitagawa2014computational}.
This queue calculation is applicable to PGMs of any partially observable Markov decision processes, not just to simple PGMs for SLAM.
Refer to \ref{apdx:pgm_slam:PGM} for the formulae of the predictive/smoothing distributions. 
In general online self-localization, 
the belief, $ bel (x_{t}) $, shown in \ref{apdx:pgm_slam:SLAM}, which is the distribution when $i=j$, is estimated without using the queue. 
{This is called filtering~\citep{kitagawa2014computational,thrun2005probabilistic}.}
The smoothing distribution can correct past self-position estimates from later observations.
SpCoSLAM 2.0~\citep{ataniguchi2020spcoslam2} introduced the fixed-lag smoother for sequential and accurate state estimation.
The predictive distribution predicts the future self-position from the current state and the learning result of the environment, providing a trajectory that avoids obstacles.
Additionally, long-term predictions can be made by repeating predictions ahead by one term.
For example, such predictions have already been realized in a map-based motion model~\citep{thrun2005probabilistic} and a stochastic model predictive control~\citep{Li2019c} for autonomous vehicles.
In addition, predictions related to the generation of spatial behavior are discussed in \ref{apdx:future:navigation}.

\section{Conclusion}
\label{sec:conclusion}

We sought to bridge the findings of HFs in neuroscience and SLAM methods in AI and robotics.
We summarized the SLAM methods in PGMs and the neuroscientific findings of the HFs and investigated their associations.
{This paper presents a case study on the framework reported by \citet{Yamakawa2021-yy}.}
The main contribution of this study is the construction of a PGM for the HF that satisfies the {evaluation} criteria for the BRA design. 
Specifically, the BIF was designed to be consistent with the anatomy of the HF, and the HCD was then designed by extending the existing SLAM model to be consistent with the structure of BIF.
{We intend to evaluate the effectiveness of this framework and HF-PGM in future studies.}
In addition, the generation-inference process allocation solved particular problems regarding the mapping of PGMs to brain circuits.
The HF-PGM is significantly different from the previous SLAM models; it integrates LEC and MEC and introduces a discrete-event queue.
Such structures were not found in most SLAMs and are very suggestive in engineering modeling.

This study operates as part of the grand challenge of realizing the whole-brain architecture using PGMs~\citep{Taniguchi2021wb-pgm}.
A whole-brain PGM will be expected to integrate submodules of PGMs for multiple brain regions by using the Neuro-SERKET architecture~\citep{Taniguchi2020neuro}. 
In the case of HF-PGM, visual information obtained from other PGM modules corresponding to its surrounding areas, including the visual cortex, could be connected as observed variables.
Other areas of HF connection, such as the prefrontal cortex, were not targeted by HF-PGM in this study.
Integration with a PGM module that mimics such areas can be considered in future studies.
{Other detailed discussions related to open questions and future perspectives are presented in \ref{apdx:future}.}

\section*{Acknowledgement}
This work was partially supported by the Japan Society for the Promotion of Science (JSPS) KAKENHI under Grant JP20K19900 and by the Ministry of Education, Culture, Sports, Science and Technology (MEXT)/JSPS KAKENHI, under Grant JP16H06569 in \#4805 (Correspondence and Fusion of Artificial Intelligence and Brain Science) and JP17H06315 in \#4905 (Brain information dynamics underlying multi-area interconnectivity and parallel processing).

We would like to thank Editage (www.editage.com) for English-language editing.

\bibliographystyle{model5-names}\biboptions{authoryear}
\bibliography{HPF-PGM}

\appendix
\setcounter{figure}{0}

\clearpage
\section{{Construction methodology for brain reference architecture}}
\label{apdx:sec:bra}

This section describes the methodology for constructing the BRA~\citep{Yamakawa2021-yy}.
Herein, the methodology used to build the BRA model (i.e., the structure-constrained interface decomposition (SCID) method) is introduced.

The whole-brain architecture approach involves iterative and incremental development with the aim of producing a system corresponding to the whole brain capable of general-purpose problem-solving. 
This approach is defined as ``to create a human-like artificial general intelligence by learning from the architecture of the entire brain.''\footnote{As the premise for this definition, the central whole-brain architecture hypothesis was set as ``the brain combines modules, each of which can be modeled with a machine-learning algorithm to attain its functionalities, thereby combining machine-learning modules such that the brain enables us to construct a generally intelligent machine with human-level or super-human cognitive capabilities.''}.
In individual projects, several tasks are solved by assigning them to partial circuits in the brain.

The SCID method is a hypothesis-building method for creating a hypothetical component diagram consistent with the neuroscientific findings. It is required to ensure consistency between neuroscientific findings and engineering feasibility to build brain-inspired models.
In current neuroscience research, findings on anatomical structures at the mesoscopic level are obtained on the near-full-brain scale, usually using rodents as the model organism.
Therefore, the SCID method can be applied to a wide area of the brain.

To construct the BRA, which includes the design information for brain-inspired AI, anatomical information is first collected, and a BIF is constructed.
A BIF is an information flow diagram that describes the mesoscopic-level anatomy of the whole brain. 
Therefore, it is not intended for a specific task in the environment.
BIF, which is a graph, is a basic structure comprising a node (i.e., circuit) and a directed link (i.e., connection).
The SCID method mainly designs HCDs that correspond to BIF.

Using the SCID method, an HCD consistent with the anatomical structure in the region of interest is obtained by the following three-step process: 
\begin{description}
\item[Step 1]
    While referring to the findings of various studies related to cognitive behaviors of humans and animals, the premise of SCID applicability is established. 
    Specifically, the three processes are performed in parallel.
    \\
    \textbf{Step 1-A}:
    The anatomical structure around the region of interest is investigated and registered as a BIF.
    \\
    \textbf{Step 1-B}:
    The existence of a component diagram that can realize the input/output is confirmed. 
    \\
    \textbf{Step 1-C}:
    The valid brain region, the region of interest, and top-level function that it carries are determined.
\item[Step 2]
    While considering the association between circuits and connections in the region of interest of the BIF, the top-level function is decomposed into detailed functions in as many conceivable patterns as possible. 
    This step enumerates candidate HCDs.
\item[Step 3]
    HCDs that are logically inconsistent according to scientific findings in various fields (e.g., neuroscience, cognitive psychology, evolution, and development) are rejected. 
    Then, the function of components and the meaning of connections of the remaining HCDs can be assigned to the BIF.
\end{description}

\clearpage
\section{Mathematical preparation of PGMs and SLAM} 
\label{apdx:pgm_slam}

As preliminary information for computational modeling, we introduce PGMs and SLAM.

\subsection{PGMs with graphical model representation}
\label{apdx:pgm_slam:PGM}

\begin{figure}[tb]
    \begin{center}
        \includegraphics[width=0.9\linewidth]{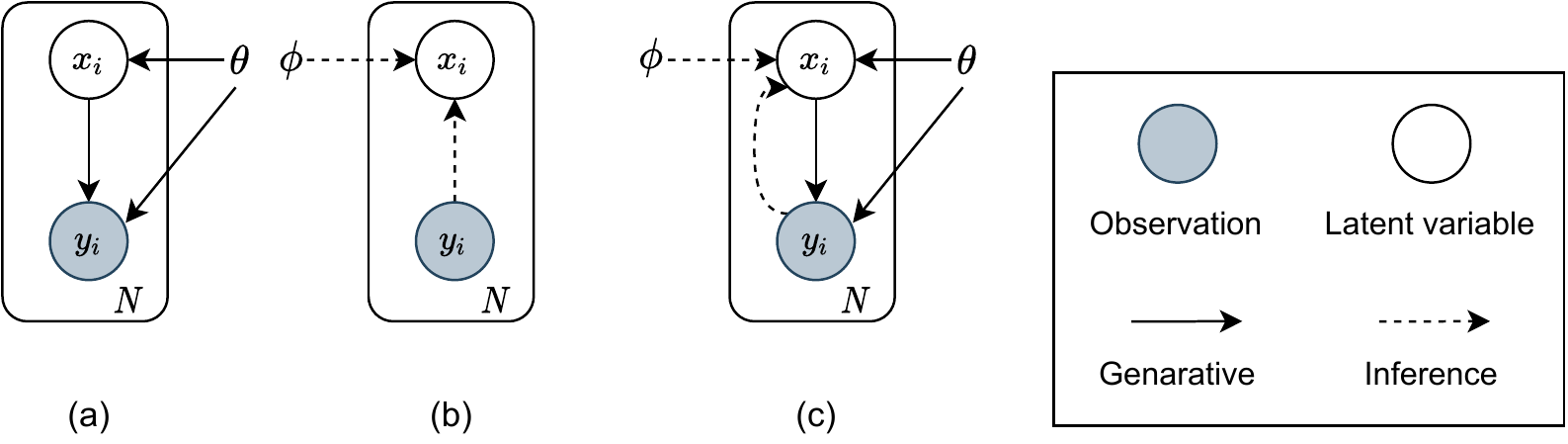}
    \end{center}
    \caption{
        Graphical model of latent variable models, such as the variational auto-encoder~\citep{kingma2014stochastic}.
        (a) Generative model, (b) inference model, and (c) both generative and inference models.
        The latent variable denotes $x_{i}$, which is a local parameter, and the observation variable denotes $y_{i}$.
        The number of data is $N$. 
        The index of the data is $i \in \{ 1, 2, \dots, N \}$.
        The global parameter for the generative model is $\theta$, and the global parameter for the inference model is $\phi$.
    }
    \label{fig:vae_gm}
\end{figure}

We provide a theoretical introduction and describe the definitions, assumptions, and constraints involved in PGMs.
PGMs represent the process that generates observations as directed acyclic graphs.
Graphical models graphically represent the dependencies among random variables in PGMs (see Fig.~\ref{fig:vae_gm}).
Generally, observable variables are represented as gray nodes, whereas unobservable variables (i.e., latent variables) are depicted as white nodes.
Global parameters determine the shape of the probability distribution, and local parameters are latent variables that correspond to individual data.
Notably, the direction of the arrows does not simply indicate the flow of signal processing; it indicates the generative process of the observation (see Fig.~\ref{fig:vae_gm} (a)).
Basically, the arrows are attached towards the observed data. 
The vicinity to the root node indicates a high-level latent representation in the brain.

As an example, a graphical model of the variational auto-encoder, which is an auto-encoding variational Bayes~\citep{kingma2014stochastic} model, is shown in Fig.~\ref{fig:vae_gm}\footnote{This figure is modified from \citet{kingma2014stochastic}.}.
Variational auto-encoder has an inference model\footnote{The inference model is sometimes called the recognition model.}, $q_{\phi}(x | y)$, which is an encoder, and a generative model, $p_{\theta}(y | x)$, which is a decoder.
The flow of signal processing and recognition in the inference model is represented by dotted arrows (see Fig.~\ref{fig:vae_gm} (b)).
When inferring latent variables, an inference model is used to calculate the posterior probability distribution of the latent variables conditioned by the observed values. 
The inference model in variational auto-encoder is constructed using amortized variational inference~\citep{Gershman2014}, which is an approach that introduces functions for efficient approximate inferencing of latent variables.
This approach leads to important models that refer to the brain structure (see Section~\ref{sec:hpf-pgm:gipa}).

PGMs are separated into the following two phases: (i) the definition of the model structures described in the generative/inference process or graphical model and (ii) the estimating/learning procedure of the posterior/predictive probability distribution and probability.
For theoretical details on PGMs, please refer to \citet{murphy2012machine}.

The state-space models shown in Fig.~\ref{fig:slam_gm} and described below are assumed to exhibit a Markov property that adds temporal transitions to latent variable models, including the variational auto-encoder.
The state-space models have three types of distributions, depending on the time difference between the state and observation variables.
The predictive, filtering, and smoothing distributions are described as follows:
\begin{eqnarray}
\text{predictive distribution} &:& p(x_t \mid y_{1:t-1}), \label{eq:pred_prob} \\ 
\text{filtering distribution}  &:& p(x_t \mid y_{1:t}),   \label{eq:filter_prob} \\ 
\text{smoothing distribution}  &:& p(x_t \mid y_{1:T}),   \label{eq:smoother_prob} 
\end{eqnarray}
where $t$ is the time of interest, $T$ is the last time, and $(1 \leqq t \leqq T)$.
The prediction can be computed from the state transition model.
The filtering distribution is typically realized using a Bayesian filter.
Smoothing is performed by post-diction, where the past state at time $t$ is updated using the observed signal in the future time, $T$.
It is achieved via data assimilation techniques, such as Kalman smoothing and fixed-lag smoothing, on state-space models~\citep{kitagawa2014computational}.

\subsection{SLAM}
\label{apdx:pgm_slam:SLAM}

\begin{figure}[tb]
    \begin{center}
        \includegraphics[width=0.9\linewidth]{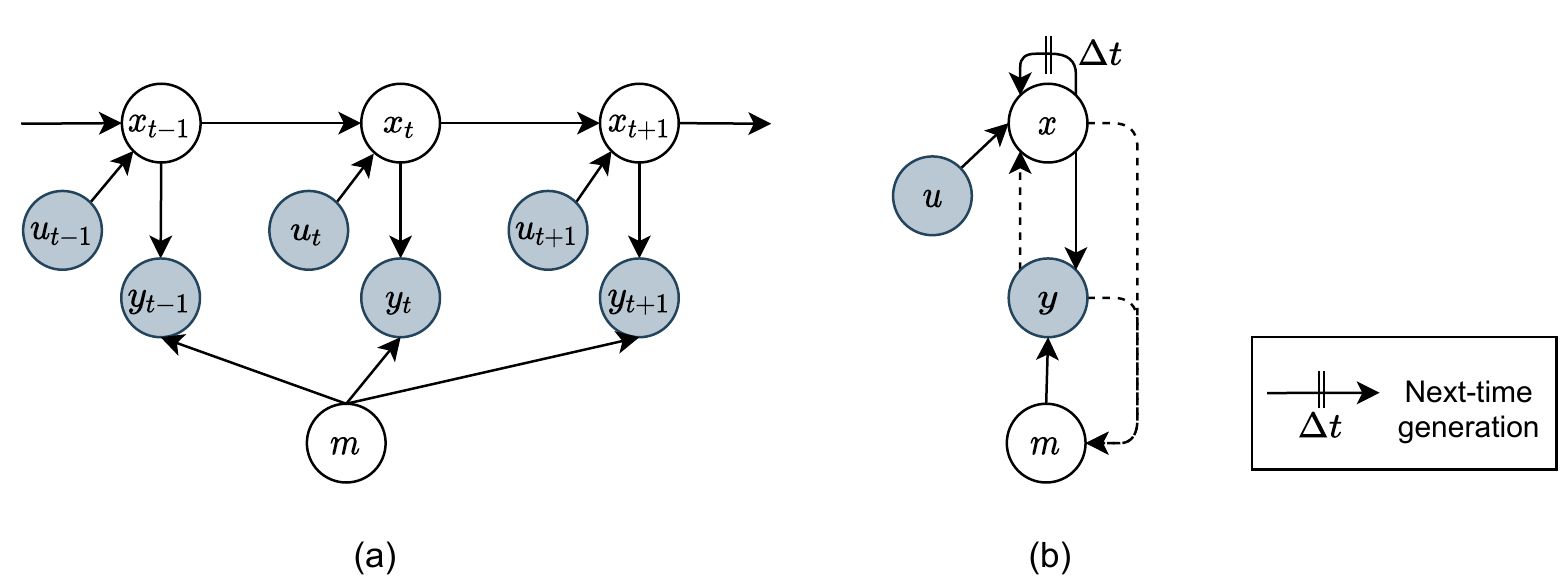}
    \end{center}
    \caption{Graphical model representations of SLAM.
    SLAM methods are represented by PGMs based on a partially observable Markov decision process.
    The self-position denotes $x_{t}$, environmental map $m$, control variable $u_{t}$, and observation variable $y_{t}$.
    Global parameters are omitted.
    (a) Typical drawing of SLAM~\citep{thrun2005probabilistic}.
    (b) Compressed drawing of time notation with inference processes, including the original notation of this paper.
    The flat arrow with $\Delta t $ indicates the generation of the variable in the next time step.
    }
    \label{fig:slam_gm}
\end{figure}

In this section, we describe the theory and classification of SLAM~\citep{thrun2005probabilistic} and existing methods based on PGMs.
SLAM is a common approach for spatial representation in robotics.
A graphical model of SLAM, representing the transition properties of the state, control, and observation, is shown in Fig.~\ref{fig:slam_gm}\footnote{In this section, because Fig.~\ref{fig:slam_gm} (b) is shown for reference, the SLAM method that introduces the inference model is not explained.
Here, to clarify that circulations with time advancing are acceptable for PGMs, the notation, ``\textit{next time generation process}'' is introduced.
The next time generation process is represented by a double line orthogonal to the generation arrow and the symbol $\Delta t$.}.
This graphical model is commonly referred to as the partially observable Markov decision process.

There are three approaches to solving the SLAM problem: Bayes filtering~\citep{montemerlo2002fastslam,gridbasedfastslam2007}, optimization by scan matching~\citep{Zhang2017}, and pose-graph optimization~\citep{Olson2006}.
In this study, we primarily focused on the Bayes filter-based online SLAM in PGMs for state-space models.
Therefore, the Bayes filtering operation is detailed below.
Scan-matching geometrically associates multiple sensor observations with one another.
Pose-graph optimization adjusts the positions based on the constraints of a graph representing the trajectory of the robot.
Furthermore, a visual SLAM~\citep{ref:taketomi2017visual} involves constructing a 3D map from images.

First, we discuss self-localization, which is a sub-problem of SLAM.
The Bayes filter is an algorithm used to estimate the posterior probability distribution of the position with respect to the entire space in the self-localization problem.
In probabilistic robotics, it is the principal algorithm for calculating beliefs; however, because it is not a practical algorithm, approximation methods (e.g., Kalman and particle filters) are applied.
Belief is a probability distribution calculated on a subjective basis that reflects the robot's internal knowledge of the state of the environment.
The belief distribution, $bel(x_{t})$, for a state, $x_{t}$, can be expressed by Equation~(\ref{eq:sinnen}). 
This belief distribution represents the posterior probability distribution on the state space conditioned on the observation, $y_{1:t}$, and the control value, $u_{1:t}$, at time $t$.
\begin{eqnarray}
bel(x_t)=p(x_t\mid y_{1:t},u_{1:t}). \label{eq:sinnen}
\end{eqnarray}

The Bayes filter is a sequential process that relies on two important assumptions: prediction and filtering.
Equation (\ref{eq:yosoku}) is the prediction by the motion model, and Equation (\ref{eq:kousin}) represents filtering by the measurement update.
This prediction is called ``dead reckoning'' in the navigation field of a voyage and calculates the position by integrating the amount of movement of the robot.
The measurement update corrects the error by observation.
An iterative update rule is applied to compute $bel(x_{t})$ from $bel(x_{t-1})$. 
The process to estimate belief is shown as follows:
\begin{eqnarray}
    \overline{bel}(x_t) &=& p(x_t\mid y_{1:t-1},u_{1:t}) \label{eq:sinnenyosoku} \nonumber \\
    &=& \int p(x_t\mid x_{t-1},u_{t})bel(x_{t-1}) dx_{t-1}, \label{eq:yosoku}
\end{eqnarray}
\begin{eqnarray}
    bel(x_t) &=& \eta p(y_t\mid x_{t})\overline{bel}(x_t). \label{eq:kousin}
\end{eqnarray}
Here, the belief distribution, $bel(x_{t-1})$, is already estimated at a previous time, $t-1$. 
The $\eta$ is the normalization term.

In addition to the aforementioned self-localization, SLAM must simultaneously estimate the environmental map.
The PGM-based SLAM methods are estimated based on Bayes filters, such as landmark-based~\citep{montemerlo2002fastslam}/grid-based FastSLAM~\citep{gridbasedfastslam2007}.
FastSLAM estimates joint posterior distribution, as shown below:
\begin{eqnarray}
    &&p(x_{0:t}, m \mid u_{1:t}, y_{1:t}) \nonumber \\
    &&=\underbrace{p(m \mid x_{0:t}, y_{1:t})}_\text{mapping}\underbrace{p(x_{0:t} \mid u_{1:t}, y_{1:t})}_\text{ self-localization },
    \label{eq:fastslam}
\end{eqnarray}
where $x_{t}$ represents the 2-dimensional (2D) coordinates in the Cartesian coordinate system and the head-direction of the agent.
$y_{t}$ represents the distance values to the obstacle obtained from a depth sensor (e.g., laser range finder) or the image features obtained from a camera.
The joint distribution of trajectory $x_{0:t}$ and map $m$ can be decomposed by factorization into two processes.
The first term represents mapping, and the second represents self-localization.
Therefore, the algorithm is a sequential iterative SLAM process.

\begin{figure}[tb]
    \begin{center}
        \includegraphics[width=1.00\linewidth]{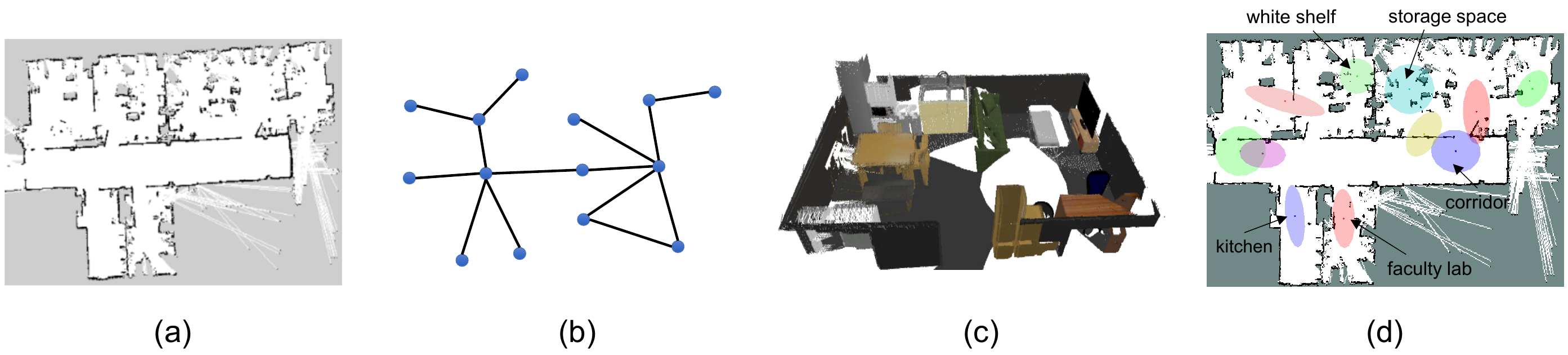}
    \end{center}
    \caption{
    Map representation examples in SLAM: 
    (a)~Occupancy grid map~\citep{gridbasedfastslam2007}. 
    (b)~Topological map~\citep{Blochliger2018},
    (c)~3D modeling map \citep{ref:labbe2014online},
    and 
    (d)~Semantic map \citep{ataniguchi2020spcoslam2}.
    }
    \label{fig:maps}
\end{figure}

Maps are constructed primarily for use in path-planning and navigation.
The map representation based on this method is shown in Figure~\ref{fig:maps}: occupancy grid map~\citep{gridbasedfastslam2007}, landmark map~\citep{montemerlo2002fastslam}, 3D modeling of the environment~\citep{ref:labbe2014online,ref:taketomi2017visual}, topological graphs~\citep{Mu2016,Blochliger2018}, and their combination~\citep{Choi2014}.
The maps can be broadly classified into metric and topological maps.
Metric maps consist of feature maps, point-cloud maps, geometric maps, occupancy grid maps, object maps, and semantic maps.
Metric maps mainly represent the presence or absence of obstacles or objects in a 2D or 3D Cartesian coordinate system.
In topological maps, nodes represent connections in places.
The graph structure comprises nodes and edges in a spatial semantic unit, such as a room, or those extracted from the metric map as post-processing. 
The combination of the metric and topological maps is referred to as a topometric map.
A semantic map is a representation that adds semantics to a metric map. More information on semantic maps is presented in Section~\ref{sec:spatial_concept}.

\clearpage
\section{{Open questions and future perspective}}
\label{apdx:future}

Herein, we discuss matters omitted in the HF-PGM alongside unresolved challenges and unclear points.

\subsection{Navigation, spatial behavior, and decision-making}
\label{apdx:future:navigation}

Spatial movement and navigation are closely related to HF functions, including spatial cognition and self-localization.
The hippocampus is a recognition unit and, therefore, does not directly issue motion commands.
Path planning and issuance of motion commands are essential for navigation, and these functions are related to regions connected to the HF~\citep{CogNeurosciSix2020}. 
In engineering, motion generation is related to reinforcement learning and path planning.

Approaches to realizing navigation in robotics are broadly divided into the following two categories.
\\
(i) \textbf{Map-based navigation}: 
One searches for a path to the goal on a map using an objective coordinate system.
It is an allocentric coordinate system navigation using SLAM and the grid map.
Classical-path planning, such as A{$^\star$} and Dijkstra's algorithms, generally belong to this type of navigation.
\\
(ii) \textbf{Visual navigation and vision-based mapless navigation}. 
The other method (i.e., visual navigation) determines the movement from an image sequence based on an egocentric viewpoint. 
This navigation is based on the subjective viewpoint signal without a coordinate system.

One could claim that (i) corresponds to the MEC, and (ii) corresponds to the LEC. 
The aforementioned two types of navigation are the same in predicting and making decisions on future states and actions based on observations. 
Approach (i) is useful for quickly searching for routes, even at distant goals, because they have an objective perspective and geometric environmental knowledge similar to those of MEC.
In contrast, many navigation types in partially observable Markov decision processes, such as reinforcement learning and vision-and-language navigation, do not use allocentric maps/coordinates, similar to LEC.
Approach (ii) enables sequential navigation based on in situ observations; however, it is assumed that it will not be suitable for long-term navigation.
Therefore, we project that MEC allocentric information processing is essential for facilitating long-term navigation.

Future prediction is expected to be effective, even when considering the generation of spatial behavior.
Humans can decide the subsequent actions in anticipation of future states. For example, one can turn right by predicting a bump into a wall if continuing straight ahead indefinitely.
Calculating the predictive value of future position and control is related to the stochastic model predictive control in the (partially observable) Markov decision process models~\citep{Li2019c}. 
Furthermore, the connection between active inference~\citep{friston2017active}, control as inference~\citep{levine2018reinforcement}, and brain functions, including the HF, is a highly interesting topic for future research.

Long-term prediction can also be formulated as probabilistic inferences by calculating predictive distributions.
\citet{ataniguchi2020spconavi} estimated the predictive distribution of future self-position and the amount of motion required from the current position.
This method is a map-based navigation based on the control as an inference~\citep{levine2018reinforcement} framework with the same PGM as in SpCoSLAM~\citep{ataniguchi_IROS2017}. 
This inference algorithm performs a forward recursive process based on dynamic programming and backward processing from the position having the maximum probability to the current position.
It realizes global path planning with speech instruction under spatial concepts acquired from the bottom-up by the robot without setting an explicit goal position.

\subsection{Connection with language and meaning}
\label{apdx:future:lang}

In this study, we constructed a model that mainly represents HF findings in rodents.
However, rodents cannot speak; therefore, this model cannot be investigated in terms of its performance in language and meaning processing.
Language processing is considered based on human neuroscientific findings.
The hippocampal declarative memory system has been identified as a potential key contributor to cognitive functions that require the online integration of multiple signal sources, such as online language processing~\citep{Duff2012}.
In the human brain, the parahippocampal place area, which encodes scenes, is active not only in visual perception but also in the auditory perception of place names~\citep{CogNeurosciSix2020}.
The parahippocampal place area is thought to be homologous to the POR in rodents. 
Therefore, this suggests that the input of place names relates to the pathway to LEC, which is responsible for abstraction in place categories.

Considering the connection between the HF and language, it is highly suggestive that robot models can learn the names of places.
These models are also closely related to semantic mapping~\citep{kostavelis2015semantic}, which grounds the names of places on a map.
Lingodroids\footnote{The Lingodroids project includes language learning by mobile robots via spatial language games to construct shared lexicons for places, distances, and directions.\\ {\url{https://itee.uq.edu.au/project/lingodroids}}} realized grounding and sharing of place nouns between mobile robots using the RatSLAM system~\citep{Schulz2011,heathlingodroids2016}.
In \citet{taguchi2011learning} and \citet{ataniguchi_IROS2017}, robots associated place nouns with specific spatial regions by receiving spoken instruction about the place names from a tutor.
\citet{gu2016learning} and \citet{Sagara2021} proposed a probabilistic model for grounding the vocabulary of spatial relative positional relationships, including the acquisition of concepts regarding distance and direction.
This type of research on the acquisition of spatial language by robots is an important topic in symbol emergence in robotics~\citep{taniguchi2019langrobo}.

Several computational models have been proposed; however, there are very few neuroscientific findings on the association between HF and language.
The acquisition of spatial language and its computational modeling are tasks that are to be continued in the future.

\subsection{Handling of time and hierarchy}
\label{apdx:future:hierarchical}

A human can supposedly handle multiple types of hierarchy~\citep{Kuipers2000}.
Hierarchies can be divided into three types: (i) temporal, (ii) spatial, and (iii) categorical/conceptual.
These hierarchies are not considered in the HF-PGM, but the introduction of a model having a hierarchy is an important direction for future studies. 
In this section, we introduce possible candidates for the hippocampal computational hierarchy model.
\\
(i) \textbf{Temporal hierarchy}. 
The brain, including the HF, can handle multiple time steps from time compression by theta-phase precession to episodic memory and temporal abstraction time-series processing. 
Time-series models, such as those in the hidden semi-Markov model~\citep{Johnson2010}, LSTM, multiple timescale RNN~\citep{Yamashita2010}, and predictive-coding-inspired variational RNN~\citep{Ahmadi2019} can be considered as models that handle multiple time transitions.
\\
(ii) \textbf{Spatial hierarchy} 
The HF has different spatial resolutions, depending on the area of the brain in the MEC, such as grid cells.
Place cells recognize places with a broader and more flexible scope than grid cells.
A model with spatial hierarchy may be reproduced using multiple resolution clustering for space, as in \citet{hagiwara2018hierarchical} and described in Section~\ref{sec:spatial_concept}. 
Topometric maps and hybrid semantic maps can also express spatial resolutions in a hierarchical manner \citep{Andrzej2017_1,Rosinol2020}.
For example, geometric structures are configured in detail on metric maps, and places are represented abstractly on a topological map. 
\\
(iii) \textbf{ Categories and conceptual hierarchy}
Relating to the two aforementioned hierarchies, hierarchy is also present in the representation of place meanings and concepts.
Hierarchical clustering using multimodal information is assumed in the computational models.
For example, the hierarchical multimodal latent Dirichlet allocation~\citep{Ando2013hMLDA}, pachinko allocation model~\citep{Wei2006}, and extended models of variational auto-encoders are applicable.
Further, the models introduced in (i) and (ii) may acquire a conceptual hierarchy as internal representations.

\subsection{Physicality and sensorimotor system}
\label{apdx:future:sensorimotor}

Anatomical/biological findings in animals other than humans and rodents are also beneficial for developing spatial cognitive systems in robotics.
For example, for a drone, it is more advantageous to imitate the brain of a flying animal (e.g., bird or bat). 
The brain structure has evolved alongside the physicality and sensory organs. 
When considering the spatial cognitive function of robots and animals, it must be noted that both the internal processes of the cognitive systems and sensorimotor organs (i.e., sensors and actuators) differ.
\\
\textbf{Actuator}:
Rodents walk on four legs, humans walk on two legs, birds fly with wings, and dolphins swim with fins.
Robots can be wheel-driven with two or four wheels, humanoid robots have two legs, and four-legged robots exist.
Drones and underwater vehicles are mainly propeller-driven.
\\
\textbf{Sensor}:
Sensors used to facilitate spatial cognition and navigation are not limited to the sensory organs of rodents and humans. 
Furthermore, some animals use ultrasonic waves and magnetic sensors for navigation.
For example, some birds use a magnetic compass to determine their alignments. 
Robots can be equipped with sensor devices, such as depth sensors, omnidirectional cameras, sonar devices, and global navigation satellite system sensors, that differ from human sensory organs.

Constructing a robot that mimics an animal can serve as a stepping stone to understanding the neural structures required for spatial cognition.
Other animals, such as birds and fish, which have similar spatial cognitive functions for constructing cognitive maps, are known to have different HF structures than mammals in evolutionary terms.
These topics are discussed in comparative hippocampal science studies~\citep{Watanabe2008}; however, we anticipate that comparisons with robot spatial cognitive models will progress in the future.
To this end, we hope that this work will inspire a comparative study of hippocampal functions between mammals and robotics-AIs.

\subsection{Mechanism for estimating the absolute speed from proximity visual stimulus}
\label{apdx:future:speed}

As mentioned in \ref{apdx:pgm_slam:SLAM}, from an engineering point of view, the absolute moving speed is measured and input to the SLAM system.
In practice, this is realistic because an accurate velocity of the signal can be obtained from mechanisms, such as the rotational speed of the wheels and the integrated value of acceleration by the gyroscope.

However, when humans are on a train, they can sense the train's speed from the scenery passing by; however, they cannot estimate this without windows. 
Hence, animals detect speed from the visual stimuli of objects that are relatively close to one another, rather than as an absolute speed obtained by engineering methods.
Within the scope of our survey, there is no knowledge that the absolute speed of movement is input to HF; the signal on the head-direction is input via the RSC, etc.

Nevertheless, the presence of neural activity associated with \textit{artificial velocity vectors} is known to occur in MEC V with grid cells~\citep{Sanders2015}.
However, it is not plausible that an artificial velocity vector can be computed within the loop of the MEC and hippocampus because the signal directly input to the MEC via POR is basically information about distant landmarks, which is not suitable for calculating an absolute speed.

However, the proposed model can successfully explain how artificial velocity vectors can be calculated using visual information about proximal objects in the HF.
First, the visual information about nearby objects arrives at the DG-CA3 via the PER and the LEC, where it is integrated with allocentric information. Subsequently, the integrated information of the proximal objects is projected to the proximal CA1 and distal Sb, where both are projected to the MEC V, allowing the artificial velocity vector to be computed using a small time-delay difference on the phase precession corresponding to one-time step deference in a discrete-event queue.
Measurement of the effect on neural activity related to artificial velocity vector on MEC V by selective inhibition in the PER-LEC pathway can be used to test this mechanism.


\clearpage
\markboth{\MakeUppercase{Appendix}}{\MakeUppercase{Appendix}}
\renewcommand*{\thesection}{Appendix \Alph{section}}

\printglossary[title={List of Abbreviations},type=acronym,style=long]

\clearpage
\forglsentries{\thislabel}{\gls{\thislabel}. }

\end{document}